\documentclass[lettersize,journal]{IEEEtran}
\usepackage{amsmath,amsfonts}
\usepackage{algorithmic}
\usepackage{algorithm}
\usepackage{array}
\usepackage[caption=false,font=normalsize,labelfont=sf,textfont=sf]{subfig}
\usepackage{textcomp}
\usepackage{graphicx}
\usepackage{hyperref}
\usepackage{tabularray}
\usepackage{amssymb}
 \usepackage{multirow}
 \usepackage{booktabs}
\usepackage{float}
\usepackage{cite}
\usepackage{url}
\usepackage{ragged2e}
\usepackage{flushend}
\usepackage[table]{xcolor} 

\usepackage{orcidlink}
\DeclareRobustCommand{\[}{\begin{equation}}
\DeclareRobustCommand{\]}{\end{equation}}
\usepackage{xcolor}
\newif\ifhighlight
\highlightfalse

\definecolor{ReviewerOne}{RGB}{0, 0, 255}
\definecolor{ReviewerTwo}{RGB}{0, 128, 0}
\definecolor{ReviewerThree}{RGB}{255, 0, 0}
\definecolor{ReviewerFour}{RGB}{255, 165, 0} 
\definecolor{ReviewerFive}{RGB}{128, 0, 128} 
\definecolor{ReviewerSix}{RGB}{75, 0, 130} 
\definecolor{ReviewerSeven}{RGB}{0, 255, 255} 

\newcommand{\revone}[1]{#1}
\newcommand{\revtwo}[1]{#1}
\newcommand{\revthree}[1]{#1}
\newcommand{\revfour}[1]{#1}

\newcommand{\revsix}[1]{#1}
\newcommand{\revseven}[1]{#1}

\definecolor{ReviewerOneB}{RGB}{0, 100, 200} 
\definecolor{ReviewerTwoB}{RGB}{0, 150, 0}   
\definecolor{ReviewerThreeB}{RGB}{200, 0, 0} 
\definecolor{ReviewerFourB}{RGB}{210, 105, 30} 
\definecolor{ReviewerFiveB}{RGB}{147, 112, 219} 
\definecolor{ReviewerSixB}{RGB}{46, 139, 87} 
\definecolor{ReviewerSevenB}{RGB}{0, 139, 139} 

\newcommand{\revoneb}[1]{\ifhighlight{\color{ReviewerOneB}#1}\else#1\fi}

\begin{document}
\title{Graph Transformers: A Survey}

\author{Ahsan Shehzad\orcidlink{0000-0003-2116-2183}, Feng Xia\orcidlink{0000-0002-8324-1859},~\IEEEmembership{Fellow,~IEEE}, Shagufta Abid\orcidlink{0009-0003-5024-8880}, Ciyuan Peng,~\IEEEmembership{Graduate Student Member,~IEEE}, Shuo Yu,~\IEEEmembership{Senior Member,~IEEE}, Dongyu Zhang, and Karin Verspoor\orcidlink{0000-0002-8661-1544}
	\IEEEcompsocitemizethanks{
		\IEEEcompsocthanksitem This work is supported by the National Natural Science Foundation of China (No. 62576075).	
		\IEEEcompsocthanksitem A.\ Shehzad and S.\ Abid are with School of Software, Dalian University of Technology, Dalian 116620, China (e-mail: \{ahsan.shehzad; shagufta.abid\}@outlook.com)
		\IEEEcompsocthanksitem F.\ Xia and K.\ Verspoor are with School of Computing Technologies, RMIT University, Melbourne, VIC 3000, Australia (e-mail: f.xia@ieee.org; karin.verspoor@rmit.edu.au)  
		\IEEEcompsocthanksitem C.\ Peng is with the Institute of Innovation, Science and Sustainability, Federation University Australia, Ballarat 3353, Australia (e-mail: ciyuan.p@ieee.org)
		\IEEEcompsocthanksitem S.\ Yu is with School of Computer Science and Technology, Dalian University of Technology, Dalian 116024, China (e-mail: shuo.yu@ieee.org)
        \IEEEcompsocthanksitem D.\ Zhang is with School of Foreign Languages and School of Software Technology, Dalian University of Technology, Dalian 116024, China (e-mail: zhangdongyu@dlut.edu.cn)        
		\IEEEcompsocthanksitem Corresponding author: Feng Xia}
}

\markboth{IEEE Transactions on Neural Networks and Learning Systems,~Vol.~0, No.~0, December~2025}%
{Shell \MakeLowercase{\textit{et al.}}: A Sample Article Using IEEEtran.cls for IEEE Journals}

\IEEEpubid{0000--0000/00\$00.00~\copyright~2025 IEEE}

\maketitle
\begin{abstract}
Graph transformers are a recent advancement in machine learning, offering a new class of neural network models for graph-structured data. The synergy between transformers and graph learning demonstrates strong performance and versatility across various graph-related tasks. This survey provides an in-depth review of recent progress and challenges in graph transformer research. We begin with foundational concepts of graphs and transformers. We then explore design perspectives of graph transformers, focusing on how they integrate graph inductive biases and graph attention mechanisms into the transformer architecture. Furthermore, we propose a taxonomy classifying graph transformers based on depth, scalability, and pre-training strategies, summarizing key principles for effective development of graph transformer models. Beyond technical analysis, we discuss the applications of graph transformer models for node-level, edge-level, and graph-level tasks, exploring their potential in other application scenarios as well. Finally, we identify remaining challenges in the field, such as scalability and efficiency, generalization and robustness, interpretability and explainability, dynamic and complex graphs, as well as data quality and diversity, charting future directions for graph transformer research.

\end{abstract}


\section{Introduction}
\IEEEPARstart{G}{raphs}, as data structures with high expressiveness, are widely used to present complex data in various domains, such as social networks, knowledge graphs, biology, chemistry, and transportation networks~\cite{velickovicEverythingConnectedGraph2023a}.
They capture both structural and semantic information from data, facilitating various tasks, such as recommendation \cite{wuGraphNeuralNetworks2023}, question answering \cite{yusufGraphNeuralNetworks2023}, anomaly detection \cite{maComprehensiveSurveyGraph2021}, sentiment analysis \cite{dasIntegratingSentimentAnalysis2024}, text generation \cite{linSurveyNeuralDatatotext2023}, and information retrieval \cite{yowMachineLearningSubgraph2023}.
\revfour{To effectively deal with graph-structured data, researchers have developed various graph learning models, such as graph neural networks (GNNs), learning meaningful representations of nodes, edges and graphs~\cite{zhangDeepLearningGraphs2020b}}. In particular, GNNs following the message-passing framework iteratively aggregate neighboring information and update node representations, leading to impressive performance on various graph-based tasks~\cite{wuComprehensiveSurveyGraph2020b}. Applications ranging from information extraction to recommender systems have benefited from GNN modelling of knowledge graphs~\cite{yeComprehensiveSurveyGraph2022}.

More recently, the graph transformer, as a new and powerful graph learning method, has attracted great attention in both the academic and industrial communities~\cite{yingTransformersReallyPerform2021e,mullerAttendingGraphTransformers2023e}. Graph transformer research is inspired by the success of transformers in natural language processing (NLP)~\cite{vaswaniAttentionAllYou2017a} and computer vision (CV)~\cite{linSurveyTransformers2022a}, coupled with the demonstrated value of GNNs. \revoneb{While GNNs rely on local message passing, which limits their ability to capture long-range dependencies and global structure, graph transformers overcome this limitation by using self-attention mechanisms that model interactions across the entire graph.} Graph transformers incorporate graph inductive bias (e.g., prior knowledge or assumptions about graph properties) to effectively process graph data~\cite{kimPureTransformersAre2022f}. Furthermore, they can adapt to dynamic and heterogeneous graphs, leveraging both node and edge features and attributes~\cite{dwivediGeneralizationTransformerNetworks2020a}. \revoneb{However, this expressiveness comes at the cost of higher computational complexity and challenges in scaling to large graphs. Thus, a key trade-off in using graph transformers lies between modeling power and computational efficiency.} Various adaptations and expansions of graph transformers have shown their superiority in tackling diverse challenges of graph learning, such as large-scale graph processing~\cite{xiaGraphLearningSurvey2021c}. Furthermore, graph transformers have been successfully employed in various domains and applications, demonstrating their effectiveness and versatility.

\IEEEpubidadjcol

Existing surveys do not adequately cover the latest advancements and comprehensive applications of graph transformers. In addition, most do not provide a systematic taxonomy of graph transformer models. For instance, Chen et al.\  \cite{chenSurveyGraphNeural2022c} focused primarily on the utilization of GNNs and graph transformers in CV, but they failed to summarize the taxonomy of graph transformer models and ignored other domains, such as NLP. Similarly, Müller et al.\ \cite{mullerAttendingGraphTransformers2023e} offered an overview of graph transformers and their theoretical properties, but they did not provide a comprehensive review of existing methods or evaluate their performance on various tasks. Lastly, Min et al. \cite{minTransformerGraphsOverview2022f} concentrated on the architectural design aspects of graph transformers, offering a systematic evaluation of different components on different graph benchmarks, but did not include significant applications of graph transformers or discuss open issues in this field.

To fill these gaps, this survey aims to present a comprehensive and systematic review of recent advancements and challenges in graph transformer research from both design and application perspectives. In comparison to existing surveys, our main contributions are as follows:
\begin{enumerate}
\item We provide a comprehensive review of the design perspectives of graph transformers, including graph inductive bias and graph attention mechanisms. We classify these techniques into different types and discuss their advantages and limitations. 
\item We present a novel taxonomy of graph transformers based on their depth, scalability, and pre-training strategy. We also provide a guide for choosing effective graph transformer architectures for different tasks and scenarios. 
\item We review the application perspectives of graph transformers in various graph learning tasks, as well as the application scenarios in other domains, such as NLP and CV tasks. 
\item We identify the crucial open issues and future directions of graph transformer research, such as the scalability, generalization, interpretability, and explainability of models, efficient temporal graph learning, and data-related issues. 
\end{enumerate}

\begin{figure*}[!ht]
\centering
\includegraphics[width=\linewidth]{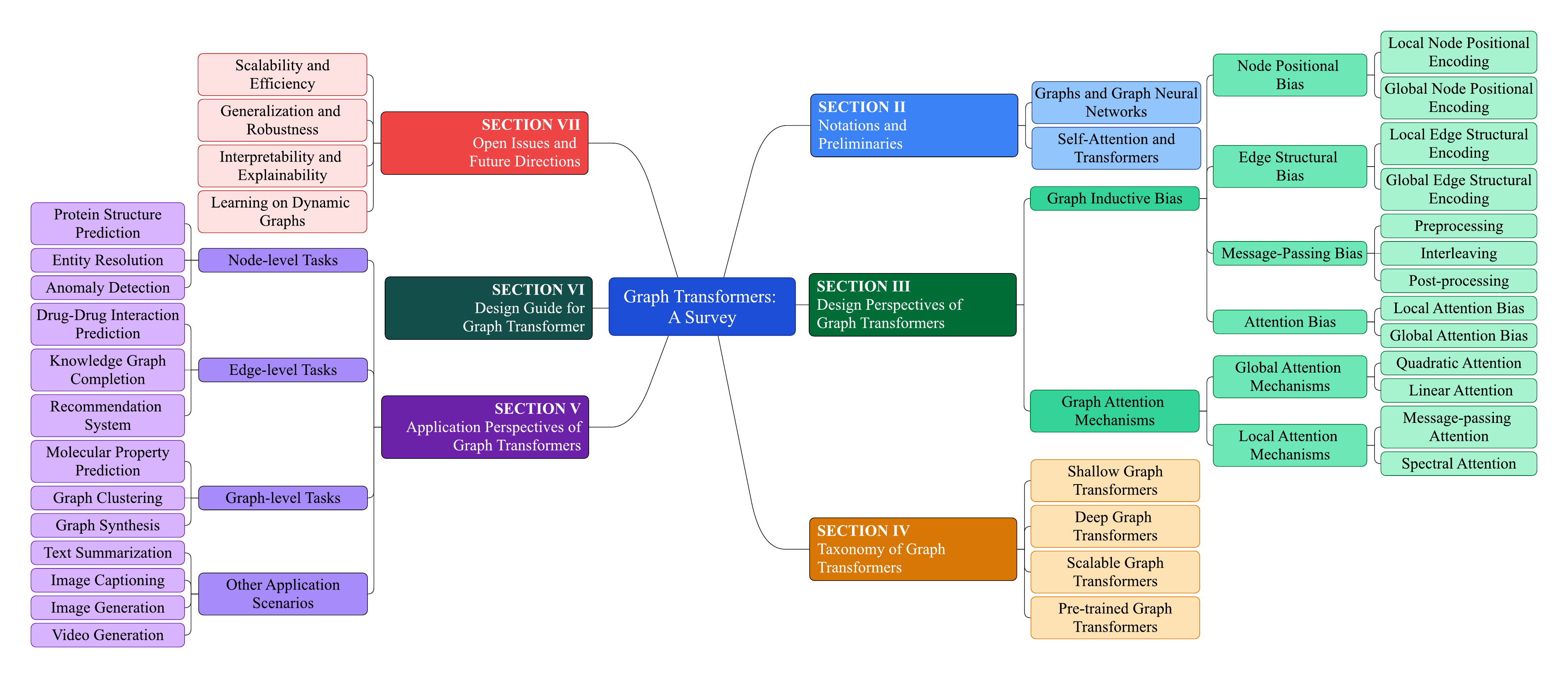}
\caption{\revone{Organization of this paper}}
\label{Fig:overview}
\end{figure*}

An overview of this paper is depicted in Figure \ref{Fig:overview}. The subsequent survey is structured as follows: Section~\ref{notations-and-preliminaries} introduces notation and preliminaries pertaining to graphs and transformers. Section~\ref{design-perspectives-of-graph-transformers} delves into the design perspectives of graph transformers that encompass graph inductive bias and graph attention mechanisms. Section~\ref{taxonomy-of-graph-transformers} presents a taxonomy of graph transformers categorizing them based on their depth, scalability and pre-training strategy. In addition, a guide is provided to select appropriate graph transformer models for various tasks and domains. Section~\ref{application-perspectives-of-graph-transformers} explores the application perspectives of graph transformers in various node-level, edge-level, and graph-level tasks, along with other application scenarios. Section~\ref{design-guide-for-effective-graph-transformers} provides a guide for designing effective graph transformer architectures for different tasks and scenarios. Section~\ref{open-issues-and-future-directions} identifies open issues and future directions for research on graph transformers. Lastly, Section~\ref{conclusion} concludes the paper and highlights its main contributions.

\section{Notations and Preliminaries}\label{notations-and-preliminaries}

In this section, we present the fundamental notations and concepts used throughout the survey. Additionally, we summarize current methods for graph learning and self-attention mechanisms, which form the foundation for graph transformers. \revoneb{The notations listed in Table \ref{table:notations} provide the mathematical language used in subsequent sections, including definitions of graphs, node and edge features, attention parameters, and positional encodings. These elements are essential to describe core mechanisms of graph neural networks (Section~\ref{design-perspectives-of-graph-transformers}), attention computation in graph transformers (Section~\ref{taxonomy-of-graph-transformers}), and task-specific architectures (Section~\ref{application-perspectives-of-graph-transformers}).}

\begin{table}[h]
\centering
\caption{Notations in this paper}
\label{table:notations}
\resizebox{\linewidth}{!}{%
\begin{tabular}{cl}
\toprule
Notation & Definition \\
\hline
$G = (V, E)$ & A graph with node set $V$ and edge set $E$ \\

$N$ & Number of nodes \\

$M$ & Number of edges \\

$\textbf{A} \in \mathbb{R}^{N \times N}$ & Adjacency matrix of graph $G$ \\
$\textbf{X} \in \mathbb{R}^{N \times d_n}$ & Node feature matrix, $\textbf{x}_i\in \textbf{X}$\\
$\textbf{F} \in \mathbb{R}^{M \times d_e}$ & Edge feature matrix\\
$\textbf{h}_v^{(l)}$ & Hidden state of node $v$ at layer $l$ \\
$\phi$ & An update function for node states \\
$\oplus$ & An aggregation function for neighbor states \\
$N(v)$ & Neighbor set of node $v$ \\
$f$ & A message function for the node and edge states \\
$\textbf{Q}, \textbf{K}, \textbf{V}$ & Query, key and value matrices for self-attention \\
$d_k$ & Dimension of query and key matrices \\
$\textbf{p}_i$ & Positional encoding of node $v_i$ \\
$d(i, j)$ & The shortest path distance between node $v_i$ and node $v_j$ \\
$\textbf{e}_{ij}$ & Edge feature between node $v_i$ and node $v_j$ \\
$a_{ij}$ & Attention score between node $v_i$ and node $v_j$ \\
$W, b$ & Learnable parameters for the self-attention layer \\
$\textbf{L} = \textbf{I}_N - \textbf{D}^{-1/2}\textbf{AD}^{-1/2}$ & Normalized graph Laplacian matrix \\

$\textbf{U}$ & Eigenvectors matrix of $\textbf{L}$ \\


\bottomrule
\end{tabular}
}
\end{table}

\subsection{Graphs and Graph Neural Networks}\label{graphs-and-graph-neural-networks}
A graph is a data structure consisting of a set of nodes (or vertices) \(V\) and a set of edges (or links) \(E\) connecting pairs of nodes. Formally, a graph is defined as \(G=(V,E)\), where \(V=\{v_1,v_2,\dots,v_N\}\) is the set of nodes with \(N\) nodes and \(E=\{e_1,e_2,\dots,e_M\}\) is the set of edges with \(M\) edges. The edge \(e_k=(v_i,v_j)\) connects node \(v_i\) and node \(v_j\), with \(i,j \in \{1,2,\dots,N\}\) and \(k \in \{1,2,\dots,M\}\). A graph can be represented by an adjacency matrix \(\textbf{A} \in \mathbb{R}^{N \times N}\), where \(A_{ij}\) indicates the presence or absence of an edge between nodes \(v_i\) and \(v_j\). Alternatively, it can be represented by the edge list \(E \in \mathbb{R}^{M \times 2}\), where each row of \(E\) contains the indices of two connected nodes. Graphs can also include node features and edge features, represented by a node feature matrix \(\textbf{X} \in \mathbb{R}^{N \times d_n}\) (where \(d_n\) is the dimension of node features) and an edge feature tensor \(\textbf{F} \in \mathbb{R}^{M \times d_e}\) (where \(d_e\) is the dimension of edge features)~\cite{chenGraphRepresentationLearning2020}.

Graph learning involves deriving low-dimensional vector representations, or embeddings, for nodes, edges, or entire graphs to capture structural and semantic information. \revoneb{For instance, embeddings represent users in social networks by encoding interactions; capture atom types and chemical bonds in molecular graphs for drug discovery; and characterize intersections and roads in navigation systems for efficient route planning.} Graph neural networks (GNNs) effectively learn from graph-structured data by propagating and aggregating information among neighboring nodes~\cite{xuHowPowerfulAre2018}. GNNs are primarily categorized into spectral and spatial methods. Spectral methods utilize graph signal processing to perform convolution operations in the spectral domain using the Fourier graph transform defined as \(\mathbf{\hat{X}}=\textbf{U}^T\textbf{X}\textbf{U}\), where \(\mathbf{U}\) contains eigenvectors of the normalized graph Laplacian \(\mathbf{L}=\mathbf{I}_N-\mathbf{D}^{-1/2}\mathbf{A}\mathbf{D}^{-1/2}\)~\cite{wangHowPowerfulAre2022a}. While spectral methods capture global graph properties, they exhibit high computational complexity and limited scalability~\cite{boSurveySpectralGraph2023a}.

Spatial methods are based on message-passing and neighborhood
aggregation, implementing convolution operations on graphs in the
spatial domain~\cite{zhangDeepLearningGraphs2020b}. The message-passing framework is defined as:
\begin{equation}\begin{aligned} \textbf{h}_v^{(l+1)}=\phi\left(\textbf{h}_v^{(l)},\bigoplus_{u \in \mathcal{N}(v)}f(\textbf{h}_u^{(l)},\textbf{h}_v^{(l)},\textbf{e}_{uv})\right) \end{aligned}, \end{equation} 
where \(\textbf{h}_v^{(l)}\) is the hidden state of node \(v\) at layer \(l\),
\(\phi\) is an update function and \(\oplus\) is an aggregation function.
\(\mathcal{N}(v)\) is the set of neighbors of node \(v\) and \(f\) is a
message function that depends on the states of the nodes and the edge features. $\textbf{e}_{uv}$ is the edge feature vector between nodes $u$ and $v$.
Spatial methods can capture local information on the graph, but
have limitations in modeling long-range dependencies, complex
interactions, and heterogeneous structures~\cite{bacciuGentleIntroductionDeep2020}.

\subsection{Self-attention and Transformers}\label{self-attention-and-transformers}
\revoneb{Self-attention is a core mechanism that enables models to relate different parts of the input sequence to each other \cite{shawSelfattentionRelativePosition2018a}. Intuitively, it helps the model decide how much attention to pay to other elements when processing a particular item. For example, in a sentence, understanding the meaning of a word often depends on its surrounding words; self-attention allows the model to capture such contextual relationships.} \revoneb{Formally, self-attention computes a weighted sum of all elements in a sequence, where the weights reflect the similarity between each element and a query vector.} It is defined as:

\begin{equation}\begin{aligned} \text{Attention}(\textbf{Q},\textbf{K},\textbf{V})=\text{softmax}\left(\frac{\textbf{QK}^T}{\sqrt{d_k}}\right)\textbf{V} \end{aligned}, \end{equation} 
where \(\textbf{Q}\), \(\textbf{K}\) and \(\textbf{V}\) are query, key and value matrices, respectively. \(d_k\) is the dimension of the query and key matrices. Self-attention can capture long-range dependencies, global context, and variable-length sequences without using recurrence or convolution. 

\begin{figure}[!ht]
\centering
\includegraphics[width=\linewidth]{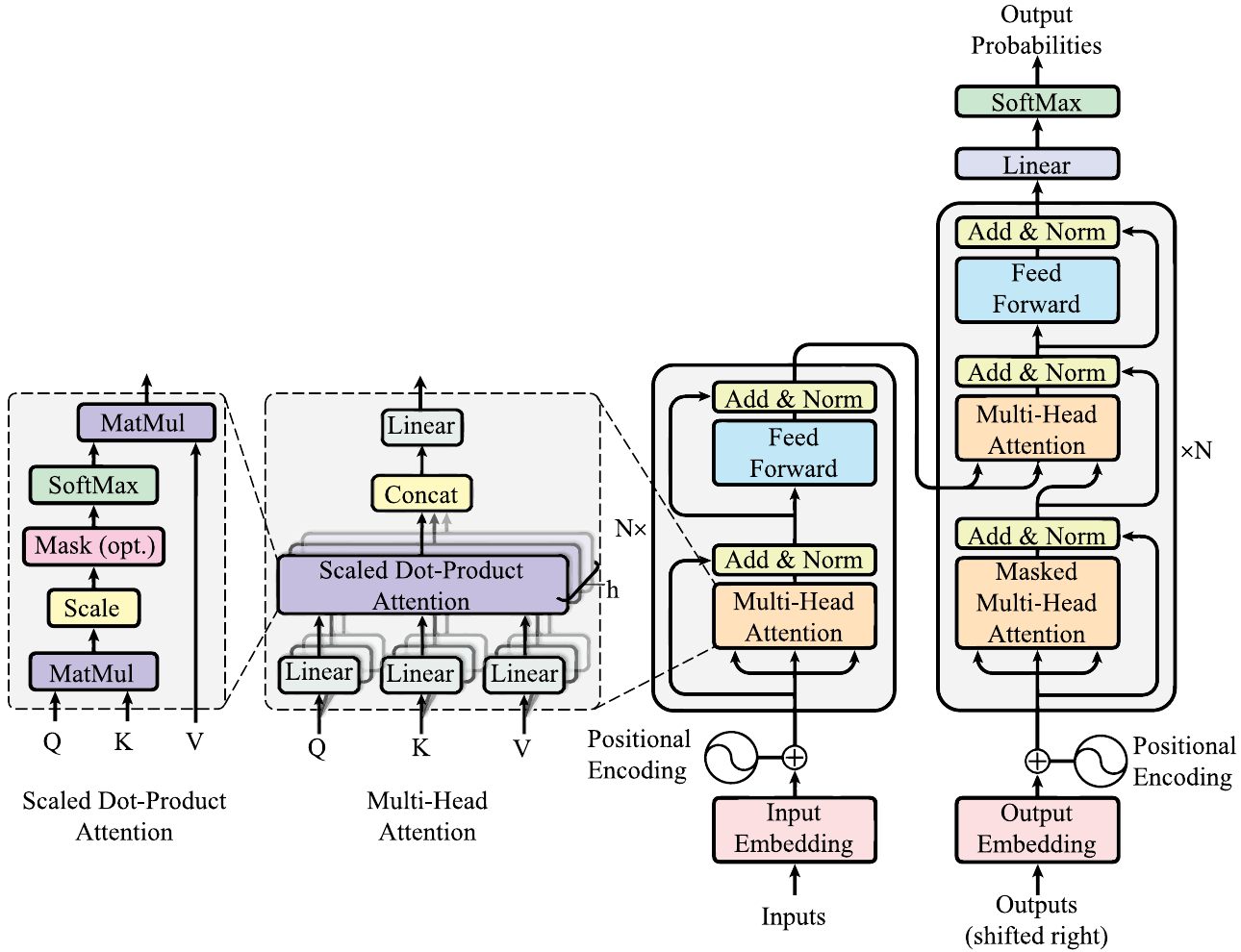}
\caption{ An illustration of transformer architecture~\cite{vaswaniAttentionAllYou2017a}.}
\label{Fig:vanilaTransformer}
\end{figure}

Transformers are neural network models that use self-attention as the main building block~\cite{vaswaniAttentionAllYou2017a}. Transformers consist of two main components: an encoder and a decoder. The encoder takes an input sequence \(\textbf{X}=\{\textbf{x}_1,\textbf{x}_2,\dots,\textbf{x}_N\}\) and generates a sequence of hidden states
\(\textbf{Z}=\{\textbf{z}_1,\textbf{z}_2,\dots,\textbf{z}_N\}\). The decoder takes an output sequence \(\textbf{Y}=\{\textbf{y}_1,\textbf{y}_2,\dots,\textbf{y}_N\}\) and generates a sequence of
hidden states \(\textbf{S}=\{\textbf{s}_1,\textbf{s}_2,\dots,\textbf{s}_N\}\). \revfour{In the final step of the decoder, these hidden states are passed through a softmax layer to compute probabilities over possible output tokens, which can then be used for downstream tasks such as sequence generation or classification.} The decoder also uses an attention mechanism to attend to the encoder's hidden states. Formally, the encoder and decoder are defined as:
\(
 \textbf{z}_i =\text{EncoderLayer}(\textbf{x}_i,\textbf{Z}_{<i}) \),  
 \(
 \textbf{s}_j =\text{DecoderLayer}(\textbf{y}_j,\textbf{S}_{<j},\textbf{Z}). 
\)
Here, \(\text{EncoderLayer}\) and \(\text{DecoderLayer}\) are composed of multiple self-attention and feed-forward sublayers. Transformers can achieve state-of-the-art results in various tasks, such as machine translation~\cite{namSurveyMultimodalBidirectional2023}, text mining~\cite{elangovan2021memorization}, document comprehension~\cite{elangovan2023effects}, image retrieval~\cite{parmarImageTransformer2018}, visual question answering~\cite{yusufGraphNeuralNetworks2023}, and image generation~\cite{sortinoTransformingImageGeneration2022}. An overview of the vanilla transformer is shown in Figure~\ref{Fig:vanilaTransformer}.

Graph transformers integrate graph inductive bias into transformers to acquire knowledge from graph-structured data~\cite{yunGraphTransformerNetworks2019d}. By employing self-attention mechanisms on nodes and edges, graph transformers can effectively capture both local and global information of the graph. \revsix{However, traditional transformer models, which are designed for sequence data, cannot be directly applied to graph data because graph-structured data does not have a natural ordering, as is the case with sequence data. In graph data, nodes and edges are interconnected in a non-sequential manner, and this necessitates the modification of the standard self-attention mechanism. To address this challenge, graph transformers modify the attention mechanism by incorporating graph-specific inductive biases such as positional encoding based on graph structure and attention over both node and edge features. These adaptations enable graph transformers to learn dependencies not only along the sequential dimensions but also in the graph topology, capturing relationships between nodes even in non-sequential configurations.} \cite{mullerAttendingGraphTransformers2023e, xia2021chief}.

\section{Design Perspectives of Graph Transformers}\label{design-perspectives-of-graph-transformers}

\revseven{In this section, we introduce the concept of “design perspectives” in Graph Transformers, referring to the core principles and strategies that guide how graph-specific inductive biases, attention mechanisms, and structural properties are integrated into the Transformer architecture. By examining these underlying design considerations, we aim to illustrate how Graph Transformers can effectively capture both local and global relationships in graph-structured data. While this section focuses on fundamental design concepts, Section IV provides a detailed taxonomy of Graph Transformer architectures, covering the diverse ways in which these design principles are operationalized to address various tasks and trade-offs in graph learning.}

This section examines how graph inductive biases and graph attention mechanisms are incorporated into graph transformer models to handle structured data effectively.

\subsection{Graph Inductive Bias }\label{graph-inductive-bias}
Unlike Euclidean data such as text and images, graph data is non-Euclidean with complex structures and no fixed order or dimensionality, making it challenging to apply standard transformers~\cite{dwivediGeneralizationTransformerNetworks2020a}. Graph transformers address this issue by integrating graph inductive bias, which encodes structural information for better generalization. We classify graph inductive bias in graph transformers into four categories: node positional bias, edge structural bias, message-passing bias, and attention bias.

\subsubsection{Node Positional Bias}\label{node-positional-encoding}

Node positional bias provides relative or absolute positional information for nodes in a graph \cite{shivNovelPositionalencoding2019}. Given a graph \(G=(V,E)\) with \(N\) nodes and \(M\) edges, each node \(v_i \in V\) has a feature vector \(\textbf{x}_i \in \mathbb{R}^d_n\). \revfour{A graph transformer aims to learn a new embedding vector \(\textbf{h}_i \in \mathbb{R}^d_k\) for each node by leveraging self-attention layers during training.} A self-attention layer is defined as:
\(
\textbf{h}_i = \sum_{j=1}^n a_{ij} W \textbf{x}_j + b,
\)
where \(a_{ij}\) is the attention score between nodes \(v_i\) and \(v_j\), measuring the relevance or similarity of their features, and \(W\) and \(b\) are learnable parameters. Because this mechanism initially lacks structural and positional information from the nodes, node positional encoding addresses these gaps by adding positional features that capture graph semantics and inductive biases~\cite{rampasekRecipeGeneralPowerful2022a}.

\emph{Local Node Positional encoding.} 
Building on the success of relative positional encoding in NLP, graph transformers adopt a similar concept for local node positional encoding. In NLP, each token receives a feature vector capturing its relative position and relationship to other words \cite{chenSimpleEffectivePositional2021a}. Similarly, graph transformers assign feature vectors to nodes based on their distance and relationships with other nodes \cite{parkGrpeRelativePositional2022a}, preserving local connectivity and neighborhood information, which benefits tasks such as node classification, link prediction, and graph generation. A frequently used approach leverages one-hot vectors representing the hop distance between a node and its neighbors \cite{yingTransformersReallyPerform2021e}:
\(
\textbf{p}_i = [I(d(i,j)=1), I(d(i,j)=2), ..., I(d(i,j)=max)],
\)
where \(d(i,j)\) is the shortest path distance between nodes \(v_i\) and \(v_j\), and \(I\) is an indicator function. The maximum hop distance is \(max\). This technique was used by Velickovic et al.\ \cite{velickovicGraphAttentionNetworks2017c} to enhance Graph Attention Networks (GATs) with position-aware self-attention. Another method involves learnable embeddings capturing the relationship between nodes \cite{maGraphAttentionNetworks2021, liu2019shifu2}, useful when a node has multiple neighbors with different edge types or labels:
\(
 \textbf{p}_i = [f(\textbf{e}_{ij_1}), f(\textbf{e}_{ij_2}), ..., f(\textbf{e}_{ij_l})],
\)
where \(\textbf{e}_{ij}\) is the edge feature between nodes \(v_i\) and \(v_j\), \(f\) is a learnable function mapping edge features to embeddings, and \(l\) is the number of neighbors considered.

Another option is to leverage graph kernels or similarity functions to measure structural similarity between nodes~\cite{parkDeformableGraphTransformer2022a, xia2021chief}:
\(
\textbf{p}_i = [K(G_i, G_{j_1}), K(G_i, G_{j_2}), ..., K(G_i, G_{j_l})],
\)
where \(G_i\) is the subgraph consisting of node \(v_i\) and its neighbors, and \(K\) is a graph kernel function measuring subgraph similarity. Mialon et al.\ \cite{mialonGraphitEncodingGraph2021a} used this idea in their GraphiT model, applying positive definite kernels as relative positional encoding. Although local node positional encoding preserves the sparsity and locality of graph structure and can improve efficiency and interpretability, it has limited ability to capture long-range dependencies or global properties of graphs, which are often required for tasks like graph matching or alignment.

\emph{Global Node Positional encoding.} 
Global node positional encoding takes inspiration from absolute positional encoding in NLP~\cite{havivTransformerLanguageModels2022}, assigning to each node a characteristic vector that represents its position within the embedding space of the graph~\cite{pengRethinkingPositionalEncoding2022}. One method for obtaining global node positional encoding leverages eigenvectors or eigenvalues of a matrix representation (e.g., adjacency or Laplacian)~\cite{rampasekRecipeGeneralPowerful2022a}:
\(
\textbf{p}_i = [u_{i1}, u_{i2}, ..., u_{ik}],
\)
where \(u_{ij}\) is the \(j\)-th component of the \(i\)-th eigenvector of the graph Laplacian. An alternative approach relies on diffusion or random walk techniques, such as personalized PageRank or heat kernel~\cite{geislerTransformersMeetDirected2023c}. In this case, the global positional encoding of a node can be represented by its random walk transition probabilities \cite{Xia2019TETCIrandom}:
\(
\textbf{p}_i = [\pi_{i1}, \pi_{i2}, ..., \pi_{iN}],
\)
where \(\pi_{ij}\) is the probability of reaching node \(v_j\) from node \(v_i\). Another common method is graph embedding or dimensionality reduction, mapping nodes into a lower-dimensional space while preserving similarity or distance \cite{huertas-garciaExploringDimensionalityReduction2023}. In such cases:
\(
\textbf{p}_i = [y_{i1}, y_{i2}, ..., y_{ik}],
\)
where \(y_{ij}\) is the \(j\)-th component of the \(i\)-th node embedding, obtained by minimizing an objective function that preserves the graph’s structure:
\(
\min_{\textbf{Y}} \sum_{i,j=1}^N \textbf{w}_{ij} \|\textbf{y}_i - \textbf{y}_j\|^2.
\)
Here, \(\textbf{w}_{ij}\) is a weight matrix reflecting node similarity or distance. The primary aim of global node positional encoding is to enhance node attribute representation by incorporating geometric and spectral graph information. This method captures long-range dependencies and overall graph characteristics, aiding tasks like graph matching and alignment. However, it can reduce sparsity and locality in graph representations, potentially affecting efficiency and interpretability.

\subsubsection{Edge Structural Bias}\label{edge-structural-encoding} 

In graph transformers, edge structural bias provides a way to extract and interpret complex graph structure information \cite{chenStructureawareTransformerGraph2022a}. It can capture diverse structural aspects, including node distances, edge types, edge directions, and local sub-structures. Studies indicate that edge structural encoding can improve the performance of graph transformers~\cite{wuNodeformerScalableGraph2022a,puEdterEdgeDetection2022,dhingraBgtnetBidirectionalGru2021}.

\emph{Local Edge Structural encoding.} 
Local edge structural encoding captures the local structure of a graph by encoding the relative position or distance between nodes~\cite{chenStructureawareTransformerGraph2022a}. Drawing on relative positional encoding in NLP and CV~\cite{hanTransformerTransformer2021}, these methods must account for the fact that graph distances can be ambiguous due to multiple paths with varying lengths or weights. GraphiT~\cite{mialonGraphitEncodingGraph2021a} introduces local edge structural encoding by using positive definite kernels to measure node similarity based on shortest path distance:
\(
k(u,v) = \exp(-\alpha d(u,v)),
\)
where \(u\) and \(v\) are nodes, \(d(u,v)\) is the shortest path distance, and \(\alpha\) controls the decay rate. This kernel modifies the self-attention score:
\(
\text{Attention}(\textbf{Q},\textbf{K},\textbf{V}) = \text{softmax}\left(\frac{\textbf{QK}^T}{\sqrt{d_k}} + k(\textbf{Q},\textbf{K})\right)\textbf{V},
\)
where \(k(\textbf{Q},\textbf{K})\) is the matrix of kernel values. EdgeBERT~\cite{tambeEdgebertSentencelevelEnergy2021} uses edge features as additional input tokens, obtained by applying a learnable function to source and target node features of each edge, then concatenating them with node features for transformer processing. The Edge-augmented Graph Transformer (EGT)~\cite{hussainGlobalSelfattentionReplacement2022a} introduces residual edge channels, matrices that store edge information for each node pair. Initialized with an adjacency matrix or shortest path matrix, these channels are updated at each transformer layer using residual connections to adjust self-attention scores:
\(
\text{Attention}(\textbf{Q},\textbf{K},\textbf{V},\textbf{R}_e) = \text{softmax}\left(\frac{\textbf{QK}^T}{\sqrt{d_k}} + \textbf{R}_e\right)\textbf{V},
\)
where \(\textbf{R}_e\) is the residual edge channel matrix. Local edge structural encoding can capture fine-grained structural information but may overlook global graph properties, and it may increase computational cost depending on the chosen kernel or encoding strategy.

\emph{Global Edge Structural encoding.} 
Global edge structural encoding aims to represent the overall structure of a graph, which is difficult due to the absence of a natural order or coordinate system \cite{zhangMultiscaleVisionLongformer2021}. GPT-GNN~\cite{huGptgnnGenerativePretraining2020} uses graph pooling and unpooling operations to encode hierarchical structures, reducing the graph size by grouping similar nodes and then restoring the original size by assigning cluster features back to individual nodes. Graphormer~\cite{yingTransformersReallyPerform2021e} applies spectral graph theory, using eigenvectors of the normalized Laplacian matrix as global positional encoding to capture global spectral features such as connectivity and community structure. Park et al.\ \cite{parkGrpeRelativePositional2022a} extended this by using singular value decomposition (SVD), employing the left singular matrix of the adjacency matrix as global positional encoding, suitable for both symmetric and asymmetric matrices. Global edge structural encoding effectively captures coarse-grained structure at the graph level but may lose detail about individual nodes or local patterns, and its success depends on the chosen encoding technique and matrix representations.

\subsubsection{Message-passing Bias}\label{message-passing-bias}

\begin{figure}
    \centering
    \includegraphics[width=1\linewidth]{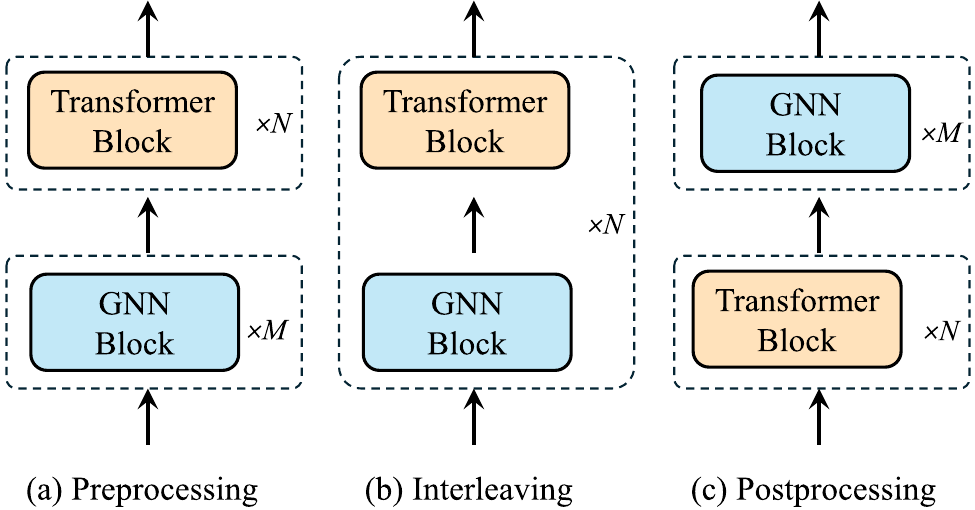}
    \caption{Incorporation of message-passing bias in transformer model.}
    \label{fig:enter-label}
\end{figure}

Message-passing bias enables graph transformers to utilize local graph structure by exchanging information between nodes and edges, addressing certain limitations of the standard transformer architecture, such as the quadratic complexity of self-attention and the lack of positional information~\cite{yinLidarbasedOnline3d2020,rampasekRecipeGeneralPowerful2022a}. Approaches to incorporate this bias in graph transformers include preprocessing, interleaving, and post-processing, each employing distinct methods of combining message-passing operations with self-attention layers (see Figure~\ref{fig:enter-label}).

\emph{Preprocessing.} Preprocessing applies message-passing operations to node features before self-attention layers \cite{yunGraphTransformerNetworks2019d}, thereby augmenting node features with local structural information. This approach retains the original transformer design and employs standard GNN-based message-passing modules. Mathematically, it is defined as:
\begin{equation}
\begin{aligned}
\textbf{h}_{v}^{(t)} &= f(\textbf{h}_{v}^{(t-1)}, \{\textbf{h}_{u}^{(t-1)}: u \in \mathcal{N}(v)\}, \{\textbf{e}_{uv}: u \in \mathcal{N}(v)\}),\\
\textbf{h}_{v}^{(t+1)} &= \text{SelfAttention}(\textbf{h}_{v}^{(t)}, \{\textbf{h}_{u}^{(t)}: u \in V\}),
\end{aligned}
\end{equation}
where \(\textbf{h}_{v}^{(0)} = \textbf{x}_{v}\). Although this method provides local structural cues, applying message-passing only once before self-attention layers may limit capturing multi-scale interactions. It can also introduce redundancy between the message-passing module and the self-attention layer.

\emph{Interleaving.} Interleaving alternates message-passing operations and self-attention layers \cite{caiGraphTransformerGraphtosequence2020a,dwivediGeneralizationTransformerNetworks2020a,ahmadGATEGraphAttention2021}, balancing local and global information processing. This arrangement supports multi-hop reasoning and can improve the expressive power of graph transformers. Mathematically, it is expressed as:
\begin{equation}
\begin{aligned}
\textbf{h}_{v}^{(t+1)} &= \theta + \text{SelfAttention}(\textbf{h}_{v}^{(t)}, \{\textbf{h}_{u}^{(t)}: u \in V\}),\\
\theta &= f(\textbf{h}_{v}^{(t)}, \{\textbf{h}_{u}^{(t)}: u \in \mathcal{N}(v)\}, \{\textbf{e}_{uv}: u \in \mathcal{N}(v)\}),
\end{aligned}
\end{equation}
where added parameters and operations can increase computational cost, and conflicting updates by message-passing and self-attention may require careful tuning.

\emph{Post-processing.} Post-processing applies message-passing to the node representations produced by self-attention layers~\cite{maGraphInductiveBiases2023b,hussainGlobalSelfattentionReplacement2022a}, refining them based on the underlying graph structure. Mathematically:
\begin{equation}
\begin{aligned}
\textbf{h}_{v}^{(t+1)} &= \text{SelfAttention}(\textbf{h}_{v}^{(t)}, \{\textbf{h}_{u}^{(t)}: u \in V\}),\\
\textbf{h}_{v}^{(T+1)} &= f(\textbf{h}_{v}^{(T)}, \{\textbf{h}_{u}^{(T)}: u \in \mathcal{N}(v)\}, \{\textbf{e}_{uv}: u \in \mathcal{N}(v)\}),
\end{aligned}
\end{equation}
where \(T\) denotes the final layer of the graph transformer. Since message-passing occurs only after self-attention, it may overlook complex interactions at earlier layers. Furthermore, overwriting information from self-attention layers can introduce inconsistencies in the final node representations.

\subsubsection{Attention Bias}\label{attention-bias}

Attention bias allows graph transformers to incorporate graph structure information into the attention mechanism without needing message-passing or positional encoding~\cite{rampasekRecipeGeneralPowerful2022a, xia2023coupled}. It modifies attention scores between nodes based on their relative positions or distances in the graph and can be categorized as either local or global, depending on whether it focuses on the local neighborhood or global topology of the graph.

\emph{Local Attention Bias.} Local attention bias restricts attention to a node's local neighborhood, similar to the message-passing mechanism in GNNs~\cite{kreuzerRethinkingGraphTransformers2021c}. It is mathematically defined as:
\begin{equation}
\alpha_{ij} = \frac{\exp(g(\mathbf{x}_i, \mathbf{x}_j) \cdot b_{ij})}{\sum_{k \in \mathcal{N}(v_i)} \exp(g(\mathbf{x}_i, \mathbf{x}_k) \cdot b_{ik})},
\end{equation}
where \(\alpha_{ij}\) is the attention score between nodes \(v_i\) and \(v_j\), \(\mathbf{x}_i\) and \(\mathbf{x}_j\) are their node features, and \(g\) computes node similarity, such as dot-product or linear transformation. The term \(b_{ij}\) is a local attention bias that adjusts the attention score based on the distance between nodes \(v_i\) and \(v_j\). This bias can be a binary mask allowing attention within a certain hop distance~\cite{nguyenDifferentiableWindowDynamic2020,velickovicGraphAttentionNetworks2017c,yunGraphTransformerNetworks2019d} or a decay function decreasing attention with increasing distance~\cite{chuTwinsRevisitingDesign2021,maGraphInductiveBiases2023b}.

\emph{Global Attention Bias.} Global attention bias integrates global topology information into the attention mechanism independent of message-passing and positional encoding \cite{hussainGlobalSelfattentionReplacement2022b}. It can be mathematically defined as follows:
\begin{equation}
\begin{aligned} 
\alpha_{ij} = \frac{\exp(g(\mathbf{x}_i, \mathbf{x}_j) + c(\mathbf{A}, \mathbf{D}, \mathbf{L}, \mathbf{P})_{ij})}{\sum_{k=1}^N\exp(g(\mathbf{x}_i, \mathbf{x}_k) + c(\mathbf{A}, \mathbf{D}, \mathbf{L}, \mathbf{P})_{ik})}.
\end{aligned} 
\end{equation} 
Here, \(c\) is the function that computes the global attention bias term, modifying attention score based on some graph-specific matrices or vectors, such as adjacency matrix \(\mathbf{A}\), degree matrix \(\mathbf{D}\), Laplacian
matrix \(\mathbf{L}\), and PageRank vector \(\mathbf{P}\). The global
attention bias term can be additive or multiplicative to the
similarity function~\cite{mialonGraphitEncodingGraph2021a,leeSelfattentionGraphPooling2019}. Generally, global attention bias can enhance the global
structure awareness and expressive power of graph transformers~\cite{leeSelfattentionGraphPooling2019}.

\subsection{Graph Attention Mechanisms}\label{graph-attention-mechanisms}

Graph attention mechanisms are widely used in constructing graph transformers \cite{soydanerAttentionMechanismNeural2022}. By dynamically weighting nodes and edges, they enable transformers to focus on the most relevant parts of a graph for a given task \cite{schwartzFactorGraphAttention2019}. Formally, a graph attention mechanism maps each node \(v_i \in V\) to a vector \(\textbf{h}_i \in \mathbb{R}^{d_k}\):
\(
\textbf{h}_i = f_n(\textbf{x}_i, \{\textbf{x}_j\}_{v_j\in \mathcal{N}(v_i)}, \{A_{ij}\}_{v_j\in \mathcal{N}(v_i)}),
\)
where \(f_n\) is a nonlinear function, \(\textbf{x}_i\) is the feature of node \(v_i\), and \(\textbf{x}_j\) is the feature of neighbor \(v_j\). The mechanism typically comprises an attention function and an aggregation function. The attention function computes a scalar weight \(\alpha_{ij}\) for each neighbor:
\(
\alpha_{ij} = \text{softmax}_i(\text{LeakyReLU}(\textbf{W}_a[\textbf{x}_i || \textbf{x}_j])),
\)
where \(\textbf{W}_a \in \mathbb{R}^{1 \times 2d_n}\) is a learnable parameter and \(||\) denotes concatenation. The aggregation function then combines weighted neighbor features:
\[
\textbf{h}_i = \textbf{W}_h\left(\textbf{x}_i + \sum_{v_j \in \mathcal{N}(v_i)}\alpha_{ij}\textbf{x}_j\right),
\]
\noindent where \(\textbf{W}_h\) is another learnable matrix. These mechanisms can also be applied to edges by treating edges as primary elements and using edge features \cite{wangEGATEdgefeaturedGraph2021}. Stacking multiple layers of graph attention allows each layer’s output to inform subsequent layers \cite{zhangGraphAttentionMultilayer2022}.

\subsubsection{Global Attention Mechanisms}\label{global-attention-mechanisms}

Global attention mechanisms define how each node allocates attention weights to all other nodes in the graph \cite{niuReviewAttentionMechanism2021}. They are commonly divided into two main categories: quadratic attention and linear attention.

\emph{Quadratic Attention.} Quadratic attention follows the standard self-attention formulation, using a softmax function on the scaled dot product of query and key vectors:
\[
\alpha_{ij} = \frac{\exp\bigl(\frac{\mathbf{q}_i^\top \mathbf{k}_j}{\sqrt{d_k}}\bigr)}{\sum_{n=1}^N \exp\bigl(\frac{\mathbf{q}_i^\top \mathbf{k}_n}{\sqrt{d_k}}\bigr)},
\]
with a computational complexity of \(O(N^2)\). Veličković et al. introduced this approach for graph transformers with multi-head attention \cite{velickovicGraphAttentionNetworks2017c}. Choromanski et al. proposed the Graph Kernel Attention Transformer (GKAT), which combines graph kernels, structural priors, and efficient attention:
\[
\begin{aligned} 
\alpha_{ij} &= \frac{\exp\bigl(\frac{\mathbf{q}_i^\top \mathbf{k}_j}{\sqrt{d_k}} + \mathbf{p}_i^\top \mathbf{p}_j\bigr)}{\sum_{n=1}^N \exp\bigl(\frac{\mathbf{q}_i^\top \mathbf{k}_n}{\sqrt{d_k}} + \mathbf{p}_i^\top \mathbf{p}_n\bigr)}, \\[6pt]
\mathbf{z}_i &= \sum_{j=1}^N \alpha_{ij} (\mathbf{U}\mathbf{V}^\top)_{ij},
\end{aligned}
\]
where \(\mathbf{p}_i\) is a positional encoding based on graph kernels, and \(\mathbf{U}, \mathbf{V}\) are low-rank matrices approximating the value matrix \cite{choromanskiBlockToeplitzMatricesDifferential2021}. Yun et al. introduced Graph Transformer Networks (GTN), which compute hierarchical representations using multiple layers of quadratic attention \cite{yunGraphTransformerNetworks2019d}. Despite their ability to capture global relationships, these mechanisms can be memory-intensive and prone to noise, and they may overlook some local information.

\emph{Linear Attention.} Linear attention mechanisms lower self-attention’s complexity from \(O(N^2)\) to \(O(N)\) by using approximations or hashing \cite{tayEfficientTransformersSurvey2022}. They typically fall into two subtypes: kernel-based and locality-sensitive. Kernel-based methods map query-key vectors into a feature space for efficient dot-product computation \cite{liuEcoformerEnergysavingAttention2022, morenoKernelMultimodalContinuous2022}. For instance, Katharopoulos et al. introduced Linear Transformer Networks (LTN) with random Fourier features \cite{katharopoulosTransformersAreRnns2020}:
\[
\begin{aligned} 
\alpha_{ij} &= \frac{\exp\bigl(\frac{\phi(\mathbf{q}_i)^\top \phi(\mathbf{k}_j)}{\sqrt{d_k}}\bigr)}{\sum_{n=1}^N \exp\bigl(\frac{\phi(\mathbf{q}_i)^\top \phi(\mathbf{k}_n)}{\sqrt{d_k}}\bigr)}, \\[6pt]
\phi(\mathbf{x}) &= \sqrt{\frac{2}{m}} \bigl[\cos(\omega_1^\top \mathbf{x} + b_1), \dots, \cos(\omega_m^\top \mathbf{x} + b_m)\bigr]^\top.
\end{aligned}
\]
Locality-sensitive linear attention uses hashing to group queries and keys into buckets for local dot-product computation \cite{leeKNNLocalAttention2022, meiImageSuperresolutionNonlocal2021, darasSmyrfefficientAttentionUsing2020}. Kitaev et al.’s Reformer model approximates self-attention with locality-sensitive hashing \cite{kitaevReformerEfficientTransformer2020}:
\[
\begin{aligned} 
\alpha_{ij} &= \frac{\exp\bigl(\tfrac{\mathbf{q}_i^\top \mathbf{k}_j}{\sqrt{d_k}}\bigr)}{\sum_{n \in B_i} \exp\bigl(\tfrac{\mathbf{q}_i^\top \mathbf{k}_n}{\sqrt{d_k}}\bigr)}, \\[6pt]
B_i &= \{j \mid h(\mathbf{q}_i) = h(\mathbf{k}_j)\},
\end{aligned} 
\]
where \(h\) is a locality-sensitive hash function. Linear attention mechanisms are more scalable for large graphs but can introduce approximation errors, hashing collisions, and partial loss of global information, potentially limiting their ability to capture complex node relationships.

\subsubsection{Local Attention Mechanisms}\label{local-attention-mechanisms}

Local attention mechanisms determine how each node computes its attention weights over a subset of nodes in the graph \cite{soydanerAttentionMechanismNeural2022a}. These mechanisms can be broadly categorized into two types: message-passing attention mechanisms and spectral attention mechanisms.

\emph{Message-passing Attention.} Message-passing attention mechanisms build on message-passing neural networks (MPNNs)~\cite{gilmerMessagePassingNeural2020}, which iteratively aggregate messages from neighboring nodes to compute node representations. These mechanisms enhance MPNNs by using self-attention to compute messages or aggregation weights \cite{yinLidarbasedOnline3d2020, liuABTMPNNAtombondTransformerbased2023, zhaoPointTransformer2021, rampasekRecipeGeneralPowerful2022a}. The computational complexity is $O(E)$, where $E$ is the number of edges. GraphSAGE~\cite{hamiltonInductiveRepresentationLearning2017a} introduced message-passing attention mechanisms for graph transformers, using mean pooling as an aggregation function and self-attention as a combination function, along with node sampling for large-scale graphs. The function is defined as:
\begin{equation}
\begin{aligned} 
\mathbf{h}_i^{(l+1)} &= \text{ReLU}(\mathbf{W}^{(l)}[\mathbf{h}_i^{(l)} \| \text{MEAN}(\{\mathbf{h}_j^{(l)}, \forall v_j \in \mathcal{N}(v_i)\})]) \\
\mathbf{z}_i &= \sum_{l=0}^L \alpha_{il} \mathbf{h}_i^{(l)}.
\end{aligned} 
\end{equation} 
\noindent Here, \(\mathbf{h}_i^{(l)}\) is the hidden vector of node \(v_i\) at layer \(l\), \(\mathbf{W}^{(l)}\) is a linear transformation matrix, \(\|\) is concatenation, and \(\text{MEAN}\) is mean pooling. \(\alpha_{il}\) is the attention weight between node \(v_i\) and layer \(l\). Javaloy et al.\ \cite{javaloyLearnableGraphConvolutional2023} proposed L-CAT, using self-attention for both aggregation and combination, and incorporating edge features with bilinear transformations. Message-passing attention mechanisms effectively preserve and leverage local graph information but face scalability issues, including limited expressiveness and dependence on graph connectivity. Their ability to capture long-range dependencies and global graph information is also limited.

\emph{Spectral Attention.}
Spectral attention mechanisms transform node features into a spectral domain, encoding graph structure with eigenvalues and eigenvectors of the graph Laplacian matrix \cite{mouLearningPayAttention2019}. They use self-attention to calculate spectral coefficients and filters \cite{boSpecformerSpectralGraph2023a}, with a computational complexity of $O(N)$. Wang et al.\ \cite{wangHowPowerfulAre2022a} proposed Graph Isomorphism Networks (GIN) as a spectral attention mechanism for graph transformers, utilizing sum pooling and self-attention. GIN integrates a graph readout scheme via set functions to encapsulate global graph characteristics, defined as:
\begin{equation}
\begin{aligned} 
\mathbf{h}_i^{(l+1)} &= \text{MLP}^{(l)}((1 + \epsilon^{(l)})\mathbf{h}_i^{(l)} + \sum_{j \in \mathcal{N}(v_i)} \mathbf{h}_j^{(l)}), \\
\mathbf{z}_i &= \sum_{l=0}^L \alpha_{il} \mathbf{h}_i^{(l)}, \\
\mathbf{z}_G &= \text{READOUT}(\{\mathbf{z}_i, \forall v_i \in V\}).
\end{aligned} 
\end{equation} 
\noindent Here, \(\text{MLP}^{(l)}\) is a multi-layer perceptron, \(\epsilon^{(l)}\) is a learnable parameter, and \(\text{READOUT}\) aggregates node output vectors into a graph output vector \(\mathbf{z}_G\). Nguyen et al.\ \cite{nguyenUniversalGraphTransformer2022c} introduced UGformer, which uses self-attention for spectral coefficient computation and incorporates edge features via bilinear transformations. Spectral attention mechanisms can incorporate graph structural information into the spectral domain, offering benefits for certain tasks. However, they face limitations such as high computational cost, memory consumption, and sensitivity to graph size and density.

\section{Taxonomy of Graph Transformers}\label{taxonomy-of-graph-transformers}

Recent years have seen a surge of interest in graph transformers. \revtwo{This section examines four key categories in the literature—shallow, deep, scalable, and pre-trained graph transformers—organized to highlight distinctions between these categories. For example, shallow and deep graph transformers are compared based on model depth and ability to capture dependencies, while scalable and pre-trained transformers focus on handling large-scale graphs and leveraging unlabeled data, respectively.} \revthree{By expanding on specific examples of Graph Transformer models in each category, we aim to elucidate the architectural choices, strengths, and limitations that guide practitioners in selecting suitable approaches for their tasks.} By analyzing representative models in each category, we aim to provide guidelines for designing effective graph transformers.

\subsection{Shallow Graph Transformers}\label{shallow-graph-transformers}

Shallow graph transformers leverage self-attention to derive node representations from graph-structured data. Inspired by transformer models that capture long-range dependencies in sequential data, these transformers extend this concept to graph data by computing self-attention weights based on both node features and graph topology \cite{zhaoAreMoreLayers2023b}. The primary goal of shallow graph transformers is to achieve high performance while minimizing computational complexity and memory usage. \revtwo{Unlike deep transformers, these models are optimized for efficiency and typically involve fewer layers, focusing on lightweight computation for smaller graphs or real-time applications.} 

These transformers can be viewed as a generalization of graph attention networks (GAT)~\cite{velickovicGraphAttentionNetworks2017c}, which use multi-head attention to compute node embeddings. However, GATs have limitations, such as the inability to model edge features and a lack of diversity among attention heads \cite{brodyHowAttentiveAre2021a}. To address these issues, several extensions have been proposed. For instance, GTN by Yun et al. \cite{yunGraphTransformerNetworks2019d} introduces edge-wise self-attention to incorporate edge information into node embeddings. Similarly, Ahmad et al. \cite{ahmadGATEGraphAttention2021} proposed the Graph Attention Transformer Encoder (GATE), which applies a masked self-attention mechanism to learn distinct attention patterns for different nodes. \revtwo{Such innovations demonstrate the adaptability of shallow graph transformers, but their limited depth constrains their ability to capture complex graph dependencies, which are better addressed by deeper models.} \revthree{In practice, shallow models often excel in lower-resource or time-sensitive scenarios due to their reduced computational overhead and simpler optimization. However, their smaller capacity may limit performance on highly complex tasks, making them most effective when graph structures are relatively simple or when inference speed is a priority.} The summary of shallow graph transformer methods is given in Table \ref{table:ShallowGraphTransformers}.

\begin{table*}[hbt!] 
\centering

\caption{Shallow graph transformers}
\label{table:ShallowGraphTransformers}
\resizebox{\linewidth}{!}{%
\rowcolors{3}{gray!10}{} 
\begin{tabular}{>{\hspace{0pt}}m{0.152\linewidth}>{\centering\hspace{0pt}}m{0.071\linewidth}>{\centering\hspace{0pt}}m{0.069\linewidth}>{\centering\hspace{0pt}}m{0.069\linewidth}>{\centering\hspace{0pt}}m{0.079\linewidth}>{\centering\hspace{0pt}}m{0.073\linewidth}>{\raggedright\hspace{0pt}}m{0.183\linewidth}>{\raggedright\hspace{0pt}}m{0.175\linewidth}>{\hspace{0pt}}m{0.038\linewidth}} 
\toprule
\multicolumn{1}{>{\Centering\hspace{0pt}}m{0.152\linewidth}}{\multirow{2}{0.152\linewidth}{\hspace{0pt}\Centering{}Model}} & \multicolumn{3}{>{\Centering\hspace{0pt}}m{0.209\linewidth}}{Graph inductive \par{}bias} & \multicolumn{2}{>{\Centering\hspace{0pt}}m{0.152\linewidth}}{Graph attention \par{}mechanisms} & \multirow{2}{0.183\linewidth}{\hspace{0pt}Application} & \multicolumn{1}{>{\Centering\hspace{0pt}}m{0.175\linewidth}}{\multirow{2}{0.175\linewidth}{\hspace{0pt}\Centering{}Datasets}} & \multirow{2}{0.038\linewidth}{\hspace{0pt}Code availability} \\ 
\cmidrule{2-4}\cmidrule(lr){5-6}
\multicolumn{1}{>{\Centering\hspace{0pt}}m{0.152\linewidth}}{} & \multicolumn{1}{>{\hspace{0pt}}m{0.071\linewidth}}{Node
  Positional encoding} & \multicolumn{1}{>{\hspace{0pt}}m{0.069\linewidth}}{Edge
  Structural Encoding} & \multicolumn{1}{>{\hspace{0pt}}m{0.069\linewidth}}{Message-Passing
  Bias} & \multicolumn{1}{>{\hspace{0pt}}m{0.079\linewidth}}{Global \par
  attention} & \multicolumn{1}{>{\hspace{0pt}}m{0.073\linewidth}}{Local \par
  Attention} &  & \multicolumn{1}{>{\Centering\hspace{0pt}}m{0.175\linewidth}}{} &  \\ 
\midrule
Graph Transformer \cite{dwivediGeneralizationTransformerNetworks2020a} & Yes (local) & Yes (local) & Yes (pre-processing) & No & Yes (message
  passing) & Graph
  representation learning & ZINC,
  PATTERN, CLUSTER & Yes \\
GAT \cite{velickovicGraphAttentionNetworks2017c} & No & No & No & Yes (linear) & Yes (message
  passing) & Node
  classification of graph-structured data & Cora,
  Citeseer, Pubmed, PPI & Yes \\
GATv2 \cite{brodyHowAttentiveAre2021a} & No & No & No & Yes (linear) & Yes (message
  passing) & Graph representation
  learning & OGB, QM9,
  VARMISUSE & Yes\par{} \par{} \\
GTN \cite{yunGraphTransformerNetworks2019d} & No & No & No & No & Yes (message
  passing) & Node
  classification on heterogeneous graphs & DBLP, ACM,
  IMDB & Yes \\
GATE \cite{ahmadGATEGraphAttention2021} & No & No & No & Yes (linear) & No & Cross-lingual
  relation and event extraction & ACE 2005
  corpus & Yes \\

Gubelmann et al. \cite{gubelmannUncoveringMoreShallow2022} & No & No & No & Yes & No & Natural
  language inference (NLI) & SNLI and MNLI & No \\
TADDY \cite{liuAnomalyDetectionDynamic2021} & Yes (global
  and local) & No & No & Yes (linear) & No & Anomaly
  detection in dynamic graphs & UCI Messages,
  Enron, DBLP, Facebook, Twitter and Reddit. & No \\
Tree Transformer \cite{shivNovelPositionalencoding2019} & Yes (local) & No & No & Yes
  (linear) & No & Program
  translation and semantic parsing & For2Lam,
  CoffeeScript-JavaScript, JOBS, GEO, ATIS & Yes \\
GAT-POS~ \cite{maGraphAttentionNetworks2021} & Yes (local) & No & No & Yes (linear) & Yes (message
  passing) & Node
  classification & Chameleon,
  Squirrel, Actor, Cora, Citeseer, Pubmed & No \\
DGT \cite{parkDeformableGraphTransformer2022a} & Yes & No & No & Yes (linear) & No & Node
  classification on graphs & Actor,
  Squirrel, Chameleon, Cora, Citeseer, twitch-gamers, ogbn-arxiv, Reddit & No \\
EDTER \cite{puEdterEdgeDetection2022} & No & No & No & Yes (linear) & Yes (message
  passing) & Edge
  detection & BSDS500,
  NYUDv2, Multicue & Yes \\
BGT-Net \cite{dhingraBgtnetBidirectionalGru2021} & No & No & No & Yes & No & Scene Graph
  Generation for Images & Visual Genome
  (VG), OpenImages (OI), Visual Relationship Detection (VRD) & No \\
PMPNet \cite{yinLidarbasedOnline3d2020} & No & No & Yes
  (pre-processing) & No & Yes (message
  passing) & 3D object
  detection & nuScenes & Yes \\
GAMLP \cite{zhangGraphAttentionMultilayer2022} & No & No & No & Yes (linear) & Yes
  (spectral) & Node
  classification & 14 real-world
  graph datasets & Yes \\
SA-GAT \cite{gaoHigherorderInteractionGoes2021} & No & No & No & Yes (linear) & Yes
  (spectral) & Graph
  classification & MUTAG, PTC, NCI1,
  NCI109, PROTEINS, ENZYMES, DD and COLLAB & No \\
GKAT \cite{choromanskiBlockToeplitzMatricesDifferential2021} & No & No & No & Yes (linear) & No & Graph
  classification and graph pattern detection & Bioinformatics
  and social networks datasets, ImageNet dataset & No \\
EcoFormer \cite{liuEcoformerEnergysavingAttention2022} & No & No & No & Yes (linear) & No & Image
  classification and long-range tasks & CIFAR-100,
  ImageNet-1K, Long Range Arena & Yes \\
Reformer \cite{kitaevReformerEfficientTransformer2020} & No & No & No & Yes (linear) & Yes (message
  passing) & NLP and
  generative tasks & Synthetic
  task, enwik8, imagenet64, WMT 2014 English-German & Yes \\
LISA \cite{soydanerAttentionMechanismNeural2022a} & No & No & No & Yes (linear) & No & Sequential
  recommendation & Alibaba,
  Amazon Video Games, MovieLens 1M, MovieLens 25M & Yes \\
\bottomrule
\end{tabular}
}
\end{table*}

\subsection{Deep Graph Transformers}\label{deep-graph-transformers}

Deep graph transformers consist of multiple stacked self-attention layers with optional skip, residual, or dense connections between layers~\cite{gongHierarchicalGraphTransformerbased2020}. They aim for higher performance through increased model depth and complexity \cite{yangTransformersOptimizationPerspective2022,Guo2024}, applying self-attention layers hierarchically to node features and graph topology. \revtwo{Unlike shallow models, these transformers can capture complex, multi-level dependencies in large and heterogeneous graphs, making them suitable for tasks requiring deep feature extraction.}

Training deeper models is challenging, often requiring techniques like PairNorm from DeeperGCN to mitigate difficulties \cite{liDeepergcnAllYou2020}. Over-smoothing can be addressed using gated residual connections and generalized convolution operators, also proposed in DeeperGCN. Other challenges, such as the loss of global attention capacity and lack of diversity among attention heads, can be tackled by approaches like DeepGraph \cite{zhaoAreMoreLayers2023b}, which uses substructure tokens and local attention to enhance focus and diversity. \revtwo{Although deep graph transformers achieve state-of-the-art performance on many graph tasks, their optimization remains sensitive to hyperparameter tuning, computational costs, and architectural choices, which require further study.} \revthree{Deeper architectures generally exhibit stronger representational power, enabling them to model complex graph relationships and long-range dependencies. Nonetheless, this comes at a higher computational cost, greater risk of vanishing gradients, and potential overfitting. Successful deep transformer designs often incorporate specialized normalization and regularization layers to preserve training stability and manage computational overhead.} Table \ref{table:DeepGraphTransformers} summarizes the methods of deep graph transformers.

\begin{table*}[hbt!] 
\centering
\caption{Deep graph transformers}
\label{table:DeepGraphTransformers}
\resizebox{\linewidth}{!}{%
\rowcolors{3}{gray!10}{} 
\begin{tabular}{>{\hspace{0pt}}m{0.152\linewidth}>{\centering\hspace{0pt}}m{0.071\linewidth}>{\centering\hspace{0pt}}m{0.069\linewidth}>{\centering\hspace{0pt}}m{0.069\linewidth}>{\centering\hspace{0pt}}m{0.079\linewidth}>{\centering\hspace{0pt}}m{0.073\linewidth}>{\raggedright\hspace{0pt}}m{0.183\linewidth}>{\raggedright\hspace{0pt}}m{0.175\linewidth}>{\hspace{0pt}}m{0.038\linewidth}} 
\toprule
\multicolumn{1}{>{\Centering\hspace{0pt}}m{0.152\linewidth}}{\multirow{2}{0.152\linewidth}{\hspace{0pt}\Centering{}Model}} & \multicolumn{3}{>{\Centering\hspace{0pt}}m{0.209\linewidth}}{Graph inductive \par{}bias} & \multicolumn{2}{>{\Centering\hspace{0pt}}m{0.152\linewidth}}{Graph attention \par{}mechanisms} & \multirow{2}{0.183\linewidth}{\hspace{0pt}Application} & \multicolumn{1}{>{\Centering\hspace{0pt}}m{0.175\linewidth}}{\multirow{2}{0.175\linewidth}{\hspace{0pt}\Centering{}Datasets}} & \multirow{2}{0.038\linewidth}{\hspace{0pt}Code availability} \\ 
\cmidrule{2-4}\cmidrule(lr){5-6}
\multicolumn{1}{>{\Centering\hspace{0pt}}m{0.152\linewidth}}{} & \multicolumn{1}{>{\hspace{0pt}}m{0.071\linewidth}}{Node
  Positional encoding} & \multicolumn{1}{>{\hspace{0pt}}m{0.069\linewidth}}{Edge
  Structural Encoding} & \multicolumn{1}{>{\hspace{0pt}}m{0.069\linewidth}}{Message-Passing
  Bias} & \multicolumn{1}{>{\hspace{0pt}}m{0.079\linewidth}}{Global \par
  attention} & \multicolumn{1}{>{\hspace{0pt}}m{0.073\linewidth}}{Local \par
  Attention} &  & \multicolumn{1}{>{\Centering\hspace{0pt}}m{0.175\linewidth}}{} &  \\ 
\midrule
DeepGraph \cite{zhaoAreMoreLayers2023b} & Yes
  (relative) & No & No & Yes (linear) & Yes (message
  passing) & Graph
  representation learning & PCQM4M-LSC,
  ZINC, CLUSTER, PATTERN & Yes \\
GraphWriter \cite{koncel-kedziorskiTextGenerationKnowledge2019c} & No & Yes & No & Yes (linear) & No & Text
  generation from knowledge graphs & AGENDA
  dataset & Yes \\
GraphLoG \cite{xuSelfsupervisedGraphlevelRepresentation2021} & No & No & Yes
  (pre-processing) & No & Yes (message
  passing) & Graph
  representation learning & ZINC15,
  MoleculeNet, Protein ego-networks & Yes \\
DIET \cite{chenSimpleEffectivePositional2021a} & Yes & No & No & Yes
  (quadratic) & No & cross-lingual
  generalization, machine translation & GLUE, XTREME,
  WMT 2018 & No \\
GRPE \cite{parkGrpeRelativePositional2022a} & Yes (local
  and global) & Yes (local
  and global) & No & Yes (linear) & No & Graph
  representation learning & ZINC,
  MolPCBA, MolHIV, PATTERN, CLUSTER, PCQM4M, PCQM4Mv2 & No \\
Graphormer \cite{yingTransformersReallyPerform2021e} & No & Yes & No & Yes
  (quadratic) & No & Graph
  representation learning & PCQM4M-LSC,
  OGB, ZINC & Yes \\
Relation-aware Self-Attention \cite{shawSelfattentionRelativePosition2018a} & No & Yes (local) & No & Yes
  (quadratic) & No & Machine
  translation & WMT 2014
  English-German and English-French & Yes \\
GraphiT \cite{mialonGraphitEncodingGraph2021a} & Yes (local and global) & No & No & Yes (quadratic) & Yes (message
  passing) & Graph
  representation learning for classification and regression tasks & MUTAG,
  PROTEINS, PTC, NCI1, ZINC & Yes \\
MagLapNet \cite{geislerTransformersMeetDirected2023c} & Yes (local
  and global) & No & No & Yes
  (quadratic) & Yes
  (spectral) & Source
  code understanding & Synthetic
  graphs, OGB Code2 & Yes \\
SAT \cite{chenStructureawareTransformerGraph2022a} & Yes (local and global) & No & No & Yes (quadratic) & Yes (message
  passing) & Node property prediction & ZINC, CLUSTER,
  PATTERN, OGBG-PPA, OGBG-CODE2 & Yes \\
Edge Transformer \cite{bergenSystematicGeneralizationEdge2021} & Yes & Yes & No & Yes (quadratic) & No & Generalization in natural language understanding and relational
  reasoning tasks & CLUTRR, CFQ
  and COGS & Yes \\
Cai et al.\ \cite{caiGraphTransformerGraphtosequence2020a} & Yes & Yes & No & Yes
  (quadratic) & No & Graph-to-sequence
  learning & LDC2015E86,
  LDC2017T10, WMT16 English-German, WMT16 English-Czech & Yes \\
SAN \cite{kreuzerRethinkingGraphTransformers2021c} & Yes (local
  and global) & No & No & Yes (quadratic) & Yes (spectral) & Graph
  representation learning & ZINC,
  PATTERN, CLUSTER, MolHIV, MolPCBA & Yes \\

ABT-MPNN \cite{ liuABTMPNNAtombondTransformerbased2023} & No & No & Yes
  (pre-processing) & Yes
  (quadratic and linear) & No & Molecular
  property prediction  & Johnson et
  al. \cite{johnson2019large}, Tox21, ClinTox, ToxCast, HIV, QM8, ESOL, FreeSolv, Lipophilicity and
  QM8. & Yes \\
UGformer \cite{nguyenUniversalGraphTransformer2022c} & No & No & No & Yes
  (quadratic) & No & Graph
  classification & COLLAB,
  IMDB-B, IMDB-M, DD, PROTEINS, MUTAG, PTC, MR, R8, R52, Ohsumed & Yes \\
DTI-GTN \cite{wangNovelMethodDrugtarget2022} & No & Yes & Yes
  (interleaving) & Yes
  (quadratic) & No & Drug-target
  interaction prediction & Peng et al.\ \cite{peng2020learning}
  dataset & No \\
Dual-GCN+Transformer \cite{dongDualGraphConvolutional2021} & No & No & No & Yes (linear) & No & Image
  Captioning & MS COCO and
  Visual Genome & Yes \\
SGTransformer \cite{sortinoTransformerbasedImageGeneration2023} & Yes & Yes & No & Yes (quadratic) & Yes (message
passing) & Scene graph
  to image generation & Visual
  Genome, COCO-Stuff and CLEVR-Dialog & Yes, \\
TPT \cite{zhangEndtoEndVideoScene2023a} & Yes & No & No & Yes & No & Video scene
  graph generation & VidHOI and
  Action Genome & No \\
\bottomrule
\end{tabular}
}
\end{table*}

\subsection{Scalable Graph Transformers}\label{scalable-graph-transformers}

Scalable graph transformers address the challenges of scalability and efficiency in applying self-attention to large-scale graphs \cite{xia2022cengcn, rampasekRecipeGeneralPowerful2022a, wuNodeformerScalableGraph2022a, wuDifformerScalableGraph2023b}. These models reduce computational cost and memory usage while maintaining or improving performance by employing techniques such as sparse attention, local attention, and low-rank approximation \cite{mullerAttendingGraphTransformers2023e, wuDifformerScalableGraph2023b}. \revtwo{Compared to deep transformers, scalable models are optimized for handling large datasets and real-world applications with limited resources.}

Several models have been proposed to enhance scalability and efficiency. Rampášek et al.\ \cite{rampasekRecipeGeneralPowerful2022a} introduced GPS, which uses low-rank matrix approximations to reduce computational complexity and achieve state-of-the-art results on diverse benchmarks. GPS decouples local real-edge aggregation from a fully-connected transformer and incorporates various positional and structural encoding to capture graph topology, offering a modular framework for local and global attention mechanisms. Cong et al.\ \cite{congDyFormerScalableDynamic2023a} developed DyFormer, a dynamic graph transformer that utilizes substructure tokens and local attention to enhance global attention diversity. \revtwo{While scalable graph transformers address key issues of memory and computation, they often involve trade-offs between scalability and expressiveness, requiring further research on optimal encoding and task-specific adaptations.} \revthree{Scalable models typically strike a balance between performance and resource demands, employing approximations or partitioning methods to handle large datasets without exhaustive computation. However, aggressive approximation can lead to reduced expressiveness, and domain-specific strategies may be needed to recover performance in certain tasks or graph structures.} For a comprehensive overview, see Table \ref{table:ScalableGraphTransformers}.

\begin{table*}[hbt!] 
\centering
\caption{Scalable graph transformers}
\label{table:ScalableGraphTransformers}

\resizebox{\linewidth}{!}{%
\rowcolors{3}{gray!10}{} 
\begin{tabular}{>{\hspace{0pt}}m{0.152\linewidth}>{\centering\hspace{0pt}}m{0.071\linewidth}>{\centering\hspace{0pt}}m{0.069\linewidth}>{\centering\hspace{0pt}}m{0.069\linewidth}>{\centering\hspace{0pt}}m{0.079\linewidth}>{\centering\hspace{0pt}}m{0.073\linewidth}>{\raggedright\hspace{0pt}}m{0.183\linewidth}>{\raggedright\hspace{0pt}}m{0.175\linewidth}>{\hspace{0pt}}m{0.038\linewidth}} 
\toprule
\multicolumn{1}{>{\Centering\hspace{0pt}}m{0.152\linewidth}}{\multirow{2}{0.152\linewidth}{\hspace{0pt}\Centering{}Model}} & \multicolumn{3}{>{\Centering\hspace{0pt}}m{0.209\linewidth}}{Graph inductive \par{}bias} & \multicolumn{2}{>{\Centering\hspace{0pt}}m{0.152\linewidth}}{Graph attention \par{}mechanisms} & \multirow{2}{0.183\linewidth}{\hspace{0pt}Application} & \multicolumn{1}{>{\Centering\hspace{0pt}}m{0.175\linewidth}}{\multirow{2}{0.175\linewidth}{\hspace{0pt}\Centering{}Datasets}} & \multirow{2}{0.038\linewidth}{\hspace{0pt}Code availability} \\ 
\cmidrule{2-4}\cmidrule(lr){5-6}
\multicolumn{1}{>{\Centering\hspace{0pt}}m{0.152\linewidth}}{} & \multicolumn{1}{>{\hspace{0pt}}m{0.071\linewidth}}{Node
  Positional encoding} & \multicolumn{1}{>{\hspace{0pt}}m{0.069\linewidth}}{Edge
  Structural Encoding} & \multicolumn{1}{>{\hspace{0pt}}m{0.069\linewidth}}{Message-Passing
  Bias} & \multicolumn{1}{>{\hspace{0pt}}m{0.079\linewidth}}{Global \par
  attention} & \multicolumn{1}{>{\hspace{0pt}}m{0.073\linewidth}}{Local \par
  Attention} &  & \multicolumn{1}{>{\Centering\hspace{0pt}}m{0.175\linewidth}}{} &  \\ 
\midrule

GPS \cite{rampasekRecipeGeneralPowerful2022a} & Yes (local
  and global) & Yes (local
  and relative) & Yes
  (interleaving) & Yes (quadratic and linear) & Yes (message
  passing) & Graph
  representation learning & ZINC,
  PCQM4Mv2, CIFAR10, MalNet-Tiny, OGB benchmarks & Yes \\
NodeFormer \cite{ wuNodeformerScalableGraph2022a} & No & No & Yes
  (relational bias) & Yes (linear) & No & Node
  classification & Citation
  networks, OGB-Proteins, Amazon2M, Mini-ImageNet, 20News-Groups & Yes \\
DIFFORMER \cite{ wuDifformerScalableGraph2023b} & No & No & No & Yes (linear) & Yes
  (spectral) & Node
  classification & Cora,
  Citeseer, Pubmed, ogbn-Proteins, Pokec, CIFAR, STL, 20News & Yes \\
DyFormer \cite{congDyFormerScalableDynamic2023a} & Yes (global) & Yes (global) & Yes
  (pre-processing) & Yes
  (quadratic) & No & Dynamic graph
  representation learning & RDS, UCI,
  Yelp, ML-10M, SNAP-Wikipedia, SNAP-Reddit & No \\
DiGress \cite{vignacDiGressDiscreteDenoising2023} & No & Yes & No & Yes & Yes & Graph
  generation & QM9, MOSES,
  GuacaMol, SBM, planar graphs & Yes \\
EXPHORMER \cite{shirzadExphormerSparseTransformers2023a} & Yes (local
  and global) & Yes (local
  and global) & Yes
  (interleaving) & Yes (linear) & Yes (spectral) & Graph
  classification & CIFAR10,
  MalNet-Tiny, MNIST, CLUSTER, PATTERN, PascalVOC-SP, COCO-SP, PCQM-Contact,
  ogbn-arxiv, & Yes \\
EGT \cite{hussainGlobalSelfattentionReplacement2022} & Yes (global) & Yes (global) & No & Yes (quadratic) & No & Node and edge
  classification & PATTERN,
  CLUSTER, TSP, MNIST, CIFAR10, ZINC, PCQM4M, PCQM4Mv2, MolPCBA and MolHIV. & Yes \\
GRIT \cite{maGraphInductiveBiases2023a} & Yes~ & No & No & Yes (quadratic) & No & Graph
  regression and classification & ZINC, MNIST,
  CIFAR10, PATTERN, CLUSTER, Peptides-func, Peptides-struct, ZINC-full,
  PCQM4Mv2 & Yes \\
HSGT \cite{zhuHierarchicalTransformerScalable2023} & No & Yes (local,
  based on shortest path distance) & No & Yes (quadratic) & Yes (message
  passing, based on neighborhood sampling) & Node
  classification & Cora,
  CiteSeer, PubMed, Amazon-Photo, ogbn-arxiv, ogbn-proteins, ogbn-products,
  Reddit, Flickr, Yelp & No \\
EGAT \cite{wangEGATEdgefeaturedGraph2021} & No & No & Yes
  (interleaving) & Yes (linear) & No & Molecular property prediction & AMLSim, Cora,
  Citeseer, PubMed & No \\

Point Transformer \cite{zhaoPointTransformer2021} & Yes
  (relative) & No & No & No & Yes (message
  passing) & 3D point
  cloud processing & S3DIS,
  ModelNet40, ShapeNetPart & Yes \\
Specformer \cite{boSpecformerSpectralGraph2023a} & No & No & No & Yes (linear) & Yes
  (spectral) & Graph
  representation learning & Synthetic,
  node-level and graph-level datasets & Yes \\
HierGAT \cite{yaoEntityResolutionHierarchical2022} & Yes (local) & No & No & Yes (linear) & No & Entity Resolution & Magellan and
  WDC product matching datasets & Yes \\

GraTransDRP \cite{chuGraphTransformerDrug2023} & Yes & No & No & Yes (linear) & No & Drug response
  prediction & GDSC & Yes \\
LGT \cite{luSimpleEfficientGraph2023a} & Yes (local and global) & No & No & Yes (linear) & Yes (message
  passing) & Molecular
  properties prediction & ZINC, QM9,
  ESOL, FreeSolv & No, \\
GTAGC \cite{hanTransformingGraphsEnhanced2023a} & Yes (global) & No & No & Yes (linear) & Yes
  (spectral) & Graph
  clustering & Citeseer,
  Cora, Pubmed & No \\
ANS-GT \cite{zhangHierarchicalGraphTransformer2022a} & Yes (local
  and global) & No & No & Yes (linear) & Yes (message
  passing) & Node classification & Cora,
  Citeseer, Pubmed, Chameleon, Actor, Squirrel, Cornell, Texas, Wisconsin & No \\
RelTR \cite{congRelTRRelationTransformer2023} & No & No & No & Yes (linear) & No & Scene graph
  generation & Visual
  Genome, Open Images V6, VRD & Yes \\
BSTG-Trans \cite{moBSTGTransBayesianSpatialTemporal2023} & Yes (local
  and global) & Yes (local
  and global) & No & Yes (linear) & Yes (message
  Passing) & Pose forecasting & Human3.6M,
  HumanEva-I, Human360K & Yes \\
GraphSum \cite{liLeveragingGraphImprove2020} & Yes (local 
  and global) & No & No & Yes (linear) & Yes (message
  passing) & Abstractive
  MDS & WikiSum and
  MultiNews & Yes \\
KG-Transformer \cite{heIntegratingGraphContextualized2019} & No & No & Yes (pre-processing) & Yes (linear) & No & Text
  generation & UMLS, PubMed,
  2010 i2b2/VA, JNLPBA, BC5CDR, GAD, EU-ADR & No \\
\bottomrule
\end{tabular}

}
\end{table*}

\subsection{Pre-trained Graph Transformers}\label{pre-trained-graph-transformers}

Pre-trained graph transformers leverage large-scale unlabeled graphs to acquire transferable node embeddings, which can be fine-tuned for downstream tasks with limited labeled data \cite{shangPretrainingGraphAugmented2019a, liGraphixt5MixingPretrained2023, rongSelfsupervisedGraphTransformer2020a}. \revtwo{These models are analogous to pre-trained language models, enabling knowledge transfer and reducing data dependence by using general-purpose embeddings.} Similar to pre-trained large language models (LLMs), these transformers are trained using self-supervised learning objectives like masked node prediction \cite{chithranandaChemBERTaLargescaleSelfsupervised2020}, edge reconstruction \cite{chenHEATHolisticEdge2022}, and graph contrastive learning \cite{leeFormNetV2MultimodalGraph2023}, capturing the inherent properties of graph data without external labels \cite{wan2023self}. The pre-trained model can then be fine-tuned on specific tasks using a smaller or domain-specific graph dataset, incorporating task-specific layers or loss functions, leading to better performance compared to training from scratch.

Challenges for pre-trained graph transformers include selecting appropriate pre-training tasks, incorporating domain knowledge, integrating heterogeneous information, and evaluating pre-training quality \cite{xiaEffectiveGeneralizableFinetuning2022}. Models like KPGT \cite{liKPGTKnowledgeguidedPretraining2022}, which uses additional domain knowledge, and KGTransformer \cite{zhangStructurePretrainingPrompt2023}, which acts as a uniform Knowledge Representation and Fusion (KRF) module, have been proposed to address these issues. \revtwo{While pre-trained models are versatile, their application is constrained by challenges related to domain-specific adaptations, evaluation consistency, and interpretability of embeddings. Further work is needed to standardize benchmarks and improve transfer learning capabilities.} \revthree{Compared to shallow or scalable transformers, pre-trained models often excel at adapting to new tasks with minimal labeled data, leveraging rich representations gleaned from large-scale training. However, this advantage can be limited if the pre-training domain diverges significantly from the target domain or if specialized graph structures require model re-parameterization.} For a comprehensive overview, see Table \ref{table:PretrainedGraphTransformers}.

\begin{table*}[hbt!] 
	\centering
	\caption{Pre-trained graph transformers}
	\label{table:PretrainedGraphTransformers}
	\resizebox{\linewidth}{!}{%
\rowcolors{3}{gray!10}{} 
\begin{tabular}{>{\hspace{0pt}}m{0.152\linewidth}>{\centering\hspace{0pt}}m{0.071\linewidth}>{\centering\hspace{0pt}}m{0.069\linewidth}>{\centering\hspace{0pt}}m{0.069\linewidth}>{\centering\hspace{0pt}}m{0.079\linewidth}>{\centering\hspace{0pt}}m{0.073\linewidth}>{\raggedright\hspace{0pt}}m{0.183\linewidth}>{\raggedright\hspace{0pt}}m{0.175\linewidth}>{\hspace{0pt}}m{0.038\linewidth}} 
\toprule
\multicolumn{1}{>{\Centering\hspace{0pt}}m{0.152\linewidth}}{\multirow{2}{0.152\linewidth}{\hspace{0pt}\Centering{}Model}} & \multicolumn{3}{>{\Centering\hspace{0pt}}m{0.209\linewidth}}{Graph inductive \par{}bias} & \multicolumn{2}{>{\Centering\hspace{0pt}}m{0.152\linewidth}}{Graph attention \par{}mechanisms} & \multirow{2}{0.183\linewidth}{\hspace{0pt}Application} & \multicolumn{1}{>{\Centering\hspace{0pt}}m{0.175\linewidth}}{\multirow{2}{0.175\linewidth}{\hspace{0pt}\Centering{}Datasets}} & \multirow{2}{0.038\linewidth}{\hspace{0pt}Code availability} \\ 
\cmidrule{2-4}\cmidrule(lr){5-6}
\multicolumn{1}{>{\Centering\hspace{0pt}}m{0.152\linewidth}}{} & \multicolumn{1}{>{\hspace{0pt}}m{0.071\linewidth}}{Node
  Positional encoding} & \multicolumn{1}{>{\hspace{0pt}}m{0.069\linewidth}}{Edge
  Structural Encoding} & \multicolumn{1}{>{\hspace{0pt}}m{0.069\linewidth}}{Message-Passing
  Bias} & \multicolumn{1}{>{\hspace{0pt}}m{0.079\linewidth}}{Global \par
  attention} & \multicolumn{1}{>{\hspace{0pt}}m{0.073\linewidth}}{Local \par
  Attention} &  & \multicolumn{1}{>{\Centering\hspace{0pt}}m{0.175\linewidth}}{} &  \\ 
\midrule
			GRAPHIX-T5 \cite{liGraphixt5MixingPretrained2023} & No & Yes & Yes (pre-processing) & Yes (linear) & No & Text-to-SQL & SPIDER, SYN,
			DK, REALISTIC, SPIDER-SSP & Yes \\
			GROVER \cite{rongSelfsupervisedGraphTransformer2020a} & No & No & Yes
			(interleaving) & Yes (linear) & Yes (message
			passing) & Molecular
			representation learning & MoleculeNet & No \\
			G-BERT \cite{shangPretrainingGraphAugmented2019a} & No & No & No & Yes (linear) & No & Medication
			recommendation & MIMIC-III & Yes \\
			ChemBERTa \cite{chithranandaChemBERTaLargescaleSelfsupervised2020} & No & No & No & Yes
			(quadratic) & No & Molecular
			property prediction & PubChem 77M,
			MoleculeNet & Yes \\
			HEAT \cite{chenHEATHolisticEdge2022} & Yes (local) & No & No & Yes (linear) & Yes (message
			passing) & Structured reconstruction & SpaceNet
			Challenge and Structured3D & Yes \\
			MPG \cite{xiaEffectiveGeneralizableFinetuning2022} & No & No & No & Yes (linear) & Yes
			(spectral) & Drug
			discovery & MoleculeNet,
			DDI, DTI & Yes \\
			LiGhT \cite{liKPGTKnowledgeguidedPretraining2022} & No & Yes (path
			encoding and distance encoding) & No & Yes (linear) & No & Molecular
			property prediction & ChEMBL29,
			BACE, BBBP, ClinTox, SIDER, Estrogen, MetStab, Tox21, ToxCast, FreeSolv,
			Lipophilicity & Yes \\
			LightGT \cite{weiLightGTLightGraph2023} & Yes (global) & No & No & Yes (linear) & No & Multimedia
			recommendation & Movielens,
			Tiktok, Kwai & Yes \\
			KGTransformer \cite{zhangStructurePretrainingPrompt2023} & No & No & No & Yes (linear) & Yes (message
			passing) & Knowledge
			graph transfer & WFC, WN18RR,
			AwA-KG, CommonsenQA & Yes \\
			CoVGT \cite{xiaoContrastiveVideoQuestion2023} & No & No & No & Yes (linear) & Yes (message-passing) & Video
			Question Answering & NExT-QA,
			TGIF-QA, TGIF-QA-R, STAR-QA, Causal-VidQA, MSRVTT-QA & Yes \\
			G-Adapter \cite{guiGAdapterStructureAwareParameterEfficient2023} & No & Yes (local) & No & Yes
			(quadratic) & No & Molecular
			graph tasks & MolHIV,
			MolPCBA, FreeSolv, ESOL, BBBP, Estrogen-$\alpha$, Estrogen-$\beta$, MetStablow,
			MetStabhigh & Yes \\
			kgTransformer \cite{liuMaskReasonPreTraining2022 } & No & No & Yes
			(interleaving) & Yes
			(quadratic) & No & Knowledge
			graph reasoning for complex logical queries & FB15k-237,
			NELL995 & Yes \\
			Li et al.\ \cite{liuMaskReasonPreTraining2022 } & Yes (local and global) & No & No & Yes (linear) & Yes (message
			passing) & Movement
			synchrony estimation & PT13 and
			Human3.6MS & No \\
			Pellegrini et al.\ \cite{pellegriniUnsupervisedPretrainingGraph2023} & No & Yes & No & Yes (linear) & No & Disease
			prediction & TADPOLE and
			MIMIC-III & Yes \\
			Video Graph Transformer (VGT) \cite{xiaoVideoGraphTransformer2022} & No & No & No & Yes (linear) & Yes (message
			passing) & Video
			Question Answering & NExT-QA,
			TGIF-QA, MSRVTT-QA & Yes \\
			Tan et al.\ \cite{tanVirtualNodeTuning2023} & No & No & No & Yes (linear) & No & node
			classification & CoraFull,
			ogbn-arxiv, CiteSeer, Cora & No \\
			GPT-GNN \cite{huGptgnnGenerativePretraining2020} & No & Yes & Yes & No & Yes & Graph
			representation learning & OAG, Amazon,
			Reddit, OAG (citation) & Yes \\
			\bottomrule
		\end{tabular}
	}
\end{table*}

\section{Application Perspectives of Graph Transformers}\label{application-perspectives-of-graph-transformers}

Graph transformers are increasingly utilized in domains involving interconnected data. This section explores their applications in graph-related tasks, categorized by analysis level: node-level, edge-level, and graph-level. Additionally, graph transformers are advancing in text, image, and video applications, where data is represented as graphs for enhanced analysis.

\subsection{Node-level Tasks}\label{node-level-tasks}

Node-level tasks involve the acquisition of node representations or the prediction of node attributes using the graph structure and node features \cite{waikhomSurveyGraphNeural2023}.

\subsubsection{Protein Structure Prediction}

In bioinformatics, graph transformers have shown significant potential in Protein Structure Prediction (PSP) \cite{tunyasuvunakoolProspectsOpportunitiesProtein2022}. Gu et al.\ \cite{guHierarchicalGraphTransformer2023} introduced HEAL, which utilizes hierarchical graph transformers on super-nodes that mimic functional motifs to interact with nodes in the protein graph, capturing structural semantics. Pepe et al.\ \cite{pepeUsingGraphTransformer2023} employed Geometric Algebra (GA) modeling to introduce a metric based on the relative orientations of amino acid residues, serving as an input feature to a graph transformer for predicting protein 3D coordinates. Chen et al.\ \cite{chenDProQGatedGraphTransformer2022a} proposed gated-graph transformers that integrate node and edge gates to regulate information flow during graph message-passing, aiding in predicting the quality of 3D protein complex structures. Despite promising results, challenges such as protein structure complexity, limited high-quality training data, and high computational demands persist \cite{chenGatedGraphTransformer2023}. Further research is needed to address these challenges and improve model accuracy.

\subsubsection{Entity Resolution}
Entity Resolution (ER) is essential in data management, aiming to identify and link disparate representations of real-world entities from various sources \cite{linEfficientEntityResolution2020}. Recent studies have shown the effectiveness of graph transformers in ER. Yao et al.\ \cite{yaoEntityResolutionHierarchical2022b} introduced Hierarchical Graph Attention Networks (HierGAT), integrating self-attention and graph attention mechanisms to leverage relationships between different ER decisions, significantly improving over traditional methods. Ying et al.\ \cite{yingTransformersReallyPerform2021i} extended the transformer architecture with structural encoding techniques to better model graph-structured data, demonstrating enhanced performance, scalability, and accuracy. Dou et al.\ \cite{douEmpoweringTransformerHybrid2022a} proposed the Hybrid Matching Knowledge for Entity Matching (GTA) method, which improves transformers for relational data by integrating hybrid matching knowledge through graph contrastive learning. This approach has shown promising results, surpassing existing entity matching frameworks. Despite challenges related to data complexity and structural information encoding, these methods have proven effective.

\subsubsection{Anomaly Detection}
Graph transformers are crucial for anomaly detection, particularly in dynamic graphs and time series data \cite{xia2023coupled, jinSurveyGraphNeural2023}. They address challenges such as encoding information for unattributed nodes and extracting discriminative knowledge from spatial-temporal dynamic graphs. Liu et al.\ \cite{liuAnomalyDetectionDynamic2021a} introduced TADDY, a transformer-based anomaly detection framework that enhances node encoding to represent structural and temporal roles in evolving graph streams. Xu et al.\ \cite{xuAnomalyTransformerTime2022} proposed the Anomaly Transformer, which uses an Anomaly-Attention mechanism to measure association discrepancy and a minimax strategy to differentiate normal and abnormal data. Chen et al.\ \cite{chenLearningGraphStructures2021} developed the GTA framework for multivariate time series anomaly detection, incorporating graph structure learning, graph convolution, and temporal dependency modeling with a transformer-based architecture. Tuli et al.\ \cite{tuliTranADDeepTransformer2022} created TranAD, a deep transformer network for anomaly detection in multivariate time series, demonstrating efficient anomaly detection and diagnosis in industrial applications. Although effective, further research is required to enhance performance and applicability across different domains \cite{Alkendi2024}.

\subsection{Edge-level Tasks}\label{edge-level-tasks}

Edge-level tasks aim to learn edge representations or predict edge
attributes based on graph structure and node features \cite{joshiEnablingAllInedge2023, shiOpenGDAGraphDomain2023}.

\subsubsection{Drug-Drug Interaction Prediction}
Graph transformers have been increasingly utilized for predicting Drug-Drug Interactions (DDIs) due to their ability to model complex drug-target relationships \cite{haubenArtificialIntelligenceData2023}. Wang et al.\ \cite{wangNovelMethodDrugtarget2022} introduced a method using a line graph with drug-protein pairs as vertices and a graph transformer network (DTI-GTN) to predict drug-target interactions. Djeddi et al.\ \cite{djeddiAdvancingDrugTarget2023a} developed DTIOG, a novel approach leveraging Knowledge Graph Embedding (KGE) and contextual information from protein sequences to predict DTIs. Despite promising results, challenges remain, such as neglecting certain intermolecular information and identifying interactions for new drugs \cite{liuImprovedDrugTarget2022}. Nonetheless, multiple studies indicate that graph transformers can effectively predict DDIs, outperforming existing algorithms \cite{al-rabeahPredictionDrugdrugInteraction2022}.

\subsubsection{Knowledge Graph Completion}
In Knowledge Graph (KG) completion, graph transformers have been extensively studied \cite{Li2024}. Chen et al.\ \cite{chenPretrainingTransformersKnowledge2023} proposed iHT, an inductive KG representation model for KG completion through large-scale pre-training, featuring an entity encoder and a neighbor-aware relational scoring function, both transformer-parameterized. This approach achieved over 25\% improvement in mean reciprocal rank compared to previous models. Liu et al.\ \cite{liuGenerativeTransformerKnowledgeGuided2023} introduced a generative transformer with knowledge-guided decoding for academic KG completion, utilizing relevant knowledge from the training corpus. Chen et al.\ \cite{chenHybridTransformerMultilevel2022} developed a hybrid transformer with multi-level fusion for multimodal KG completion, integrating visual and textual representations via coarse-grained prefix-guided interaction and fine-grained correlation-aware fusion modules.

\subsubsection{Recommendation Systems}
Graph transformers have been effectively employed in recommender systems by integrating generative self-supervised learning with a graph transformer architecture \cite{yuSelfsupervisedLearningRecommender2023a}. Xia et al.\ \cite{xiaSelfSupervisedHypergraphTransformer2022} utilized generative self-supervised learning to extract representations unsupervisedly and applied graph transformers to capture complex user-item relationships. Li et al.\ \cite{liGraphTransformerRecommendation2023a} introduced GFormer, which uses rationale-aware generative self-supervised learning to identify informative patterns in user-item interactions. This method automates self-supervision augmentation and preserves user-item relationships via collaborative rationale discovery. Despite challenges in graph construction, network design, model optimization, computation efficiency, and handling diverse user behaviors, experiments demonstrate that this approach consistently outperforms baseline models across various datasets \cite{Wang2024}.

\subsection{Graph-level Tasks}

Graph-level tasks aim to learn graph representations or predict graph
attributes based on graph structure and node features.

\subsubsection{Molecular Property Prediction}
Graph transformers are highly effective for molecular property prediction, leveraging molecular graph structures to capture essential structural and semantic information \cite{shenMolecularPropertyPrediction2019}. Chen et al.\ \cite{chenAlgebraicGraphassistedBidirectional2021a} introduced the Algebraic Graph-Assisted Bidirectional Transformer (AGBT) framework, which integrates 3D molecular information into graph invariants, addressing the lack of stereochemical information in some models. Li et al.\ \cite{liKPGTKnowledgeguidedPretraining2022} utilized Knowledge-Guided Pre-training of Graph Transformers (KPGT) in a self-supervised learning framework, emphasizing chemical bonds and modeling molecular graph structures. Buterez et al.\ \cite{buterezTransferLearningGraph2024} proposed transfer learning with graph transformers to improve molecular property prediction on sparse, high-fidelity data.

\subsubsection{Graph Clustering}
Graph transformers are increasingly employed in Graph Clustering, offering innovative methods and addressing significant challenges \cite{Liang2024}. Yun et al.\ \cite{yunGraphTransformerNetworks2019f} introduced a graph transformer network to generate new graph structures, identifying useful connections between previously unconnected nodes while learning effective node representations in an end-to-end manner. Gao et al.\ \cite{gaoPatchGTTransformerNontrainable2022a} proposed the Patch Graph Transformer (PatchGT), which segments a graph into patches via spectral clustering without trainable parameters, using GNN layers to learn patch-level representations and transformers for graph-level representations. These methodologies overcome limitations of local attention mechanisms and difficulties in learning high-level information, resulting in enhanced graph representation, improved model performance, and effective node representation.

\subsubsection{Graph Synthesis}
Graph transformers have advanced graph synthesis by enhancing graph data mining and representation learning. Traditional graph transformers with Positional encoding face limitations in node classification tasks on complex graphs due to inadequate capture of local node properties \cite{dwivediGeneralizationTransformerNetworks2020a}. To address this, Ma et al.\ \cite{maRethinkingStructuralencoding2023a} introduced the Adaptive Graph Transformer (AGT), which effectively and efficiently extracts structural patterns from graphs by learning from centrality and subgraph views, achieving state-of-the-art performance on real-world web graphs and synthetic graphs with heterophily and noise. Jiang et al.\ \cite{jiangAGFormerEfficientGraph2023a} proposed the Anchor Graph Transformer (AGFormer), using an anchor graph model for efficient and robust node-to-node message-passing, overcoming computational costs and sensitivity to graph noise. Zhu et al.\ \cite{zhuHierarchicalTransformerScalable2023} developed the Hierarchical Scalable Graph Transformer (HSGT), scaling the transformer architecture to node representation learning tasks on large-scale graphs through graph hierarchies and sampling-based training methods.
\revtwo{
\subsection{Other Application Scenarios}\label{other-application-scenarios}
Graph transformers have a wide range of applications beyond graph-structured data. They can also be utilized in scenarios involving text, images, or videos. \cite{chenSurveyGraphNeural2022d}.

\subsubsection{Text Summarization}
Text summarization, a crucial NLP task, has been significantly advanced by Graph transformers \cite{el-kassasAutomaticTextSummarization2021}. These models employ extractive, abstractive, and hybrid methods for summarization \cite{Chen2024}. Extractive summarization selects key sentences or phrases from the original text \cite{guptaAutomatedNewsSummarization2022}, while abstractive summarization interprets core concepts to produce concise summaries. Hybrid summarization combines both approaches \cite{yangRecentProgressText2023}. Despite challenges in text comprehension, main idea identification, and coherent summary generation, graph transformers have shown promising results in summary quality and efficiency \cite{kumarAbstractiveTextSummarization2023}.

\subsubsection{Image Captioning}
Graph transformers have become a powerful tool in image captioning, providing structured image representations and efficiently generating descriptive captions \cite{hossainComprehensiveSurveyDeep2019,Zhang2024}. Techniques like Transforming Scene Graphs (TSG) use multi-head attention to design graph neural networks that embed scene graphs, capturing extensive knowledge to aid word generation across various parts of speech \cite{yangTransformingVisualScene2023a}. Despite challenges like training complexity, lack of contextual information, and fine-grained details, graph transformers have shown promising results, improving sentence quality and achieving state-of-the-art performance in image captioning \cite{wangEndtoendTransformerBased2022}.

\subsubsection{Image Generation}
Graph transformers have been effectively utilized in image generation, as shown by various studies \cite{zelaszczykTexttoImageCrossModalGeneration2024}. Sortino et al.\ \cite{sortinoTransformingImageGeneration2022} proposed a transformer-based method conditioned on scene graphs, employing a decoder for sequential image composition. Zhang et al.\ \cite{zhangStyleswinTransformerbasedGan2022} introduced StyleSwin, using transformers in a generative adversarial network for high-resolution image creation. Despite challenges like redundant interactions and complex architectures, these studies have shown promising results in image quality and variety \cite{congRelTRRelationTransformer2023,liSgtrEndtoendScene2022}.

\subsubsection{Video Generation}
Graph transformers have been extensively applied in video generation. Xiao et al.\ \cite{xiaoVideoGraphTransformer2022b} introduced the Video Graph Transformer (VGT) model, using a dynamic graph transformer to encode videos by capturing visual objects, relationships, and dynamics, incorporating disentangled video and text transformers for relevance comparison. Wu et al.\ \cite{wuGenerativeVideoTransformer2021a} proposed the Object-Centric Video Transformer (OCVT), employing an object-centric approach to tokenize scenes for a generative video transformer, understanding complex spatiotemporal dynamics. Yan et al.\ \cite{yanVideoGPTVideoGeneration2021a} developed VideoGPT, which uses 3D convolutions and axial self-attention to learn downsampled discrete latent representations of raw videos, modeling these latents with a GPT-like architecture. Tulyakov et al.\ \cite{tulyakovMocoganDecomposingMotion2018} proposed MoCoGAN, mapping random vectors to video frame sequences. Despite challenges in capturing complex spatiotemporal dynamics, these methodologies have shown promising outcomes in various video generation tasks, from question answering to video summarization.}

\section{Design Guide for Graph Transformers}\label{design-guide-for-effective-graph-transformers}
Developing effective graph transformers requires meticulous attention to graph-specific challenges and careful consideration of design choices. This guide outlines practical principles for designing graph transformers across various graph types and tasks.

\begin{itemize}
\item
  \textit{Choose the appropriate type of graph transformer based on your graph data and tasks.} \revone{Selecting a graph transformer requires a detailed understanding of the graph's structure and scale. For "simple, small" graphs, characterized by fewer than $10^3$ nodes and sparse connectivity (e.g., edge-to-node ratio $<10$), shallow transformers with fewer layers (e.g., 2–4 layers) may be sufficient to capture structural information efficiently. In contrast, "complex, large" graphs, defined as having more than $10^5$ nodes, dense connectivity (edge-to-node ratio $\geq 10$), or rich attributes (e.g., temporal, weighted, or multi-relational edges), benefit from deeper models (e.g., 6–12 layers) capable of learning expressive representations. For dynamic or streaming graphs, where node or edge data evolves over time, scalable transformers leveraging techniques like subgraph sampling or incremental updates are essential \cite{Lanciano2024}. Sparse or noisy graphs, such as those with missing edges or incomplete attributes, are effectively handled by pre-trained graph transformers, which leverage knowledge transfer and task-specific fine-tuning to improve robustness.}

\item
  \textit{Design suitable structural and positional encodings for your graph data.} \revone{Effective encoding of structural and positional information is crucial for transformers to capture graph-specific inductive biases. Techniques such as Laplacian eigenvector-based encodings, spectral embeddings, or random walk-based encodings (e.g., Node2Vec or DeepWalk embeddings) can encode global and local graph structure \cite{Cai2018}. For weighted graphs, edge weights can be integrated into positional encodings using distance-weighted features or flow-based encodings. For directed graphs, direction-sensitive encodings such as signed Laplacian eigenvectors or edge direction indicators can capture asymmetry \cite{Wang2023}. Incorporating these encodings into node or edge embeddings preserves structural informativeness and supports tasks such as graph classification and link prediction.}

\item
  \textit{Optimize the self-attention mechanism for your graph data.} \revone{The self-attention mechanism in graph transformers computes pairwise attention scores, which are computationally expensive for large graphs \cite{Soydaner2022}. To address this, various optimization strategies can be employed:
  \begin{itemize}
    \item \textbf{Sampling:} Node sampling (e.g., GraphSAGE~\cite{hamiltonInductiveRepresentationLearning2017a}) or edge sampling (e.g., DropEdge~\cite{rong2020dropedge}) reduces computation by operating on representative graph subsets while preserving global properties.
    \item \textbf{Sparsification:} Techniques like top-$k$ sparsification retain only the most informative attention scores, reducing dense computations \cite{Hashemi2024}.
    \item \textbf{Partitioning:} Graph partitioning methods (e.g., METIS \cite{Karypis1998}) divide the graph into smaller, manageable subgraphs processed independently, improving scalability \cite{atalyrek2023}.
    \item \textbf{Hashing:} Methods such as locality-sensitive hashing approximate attention computations, enabling efficient processing of large node sets \cite{Peng2018}.
    \item \textbf{Regularization and normalization:} Regularizers like Dropout and attention normalization techniques mitigate overfitting and stabilize training, addressing issues like oversmoothing and oversquashing in deep transformers.
  \end{itemize}
  Each technique has trade-offs in terms of computational overhead, scalability, and performance fidelity, making the choice context-dependent.}

\item
  \textit{Utilize pre-training techniques to enhance performance.} \revone{Pre-training techniques leverage large-scale datasets or pre-trained models to improve task-specific performance. Methods such as graph contrastive learning (e.g., GraphCL \cite{You2020GraphCL}) enhance feature extraction by maximizing agreement between augmented graph views. Fine-tuning techniques adapt pre-trained models for domain-specific tasks, while distillation transfers knowledge from large, resource-intensive models to smaller, efficient ones. Pre-training is particularly beneficial for low-resource scenarios, where labeled data is scarce, enabling effective downstream task performance with limited additional training.}
\end{itemize}

\section{Open Issues and Future Directions}\label{open-issues-and-future-directions}

Despite their immense potential for learning from graph-structured data, graph transformers still face open issues and challenges that require further exploration. Here we highlight some of these open challenges.

\subsection{Scalability and Efficiency}\label{scalability-and-efficiency}

The scalability and efficiency of graph transformers face significant challenges due to high memory and computational demands, particularly with global attention mechanisms for large-scale graphs~\cite{sun2022attributed, xia2022cengcn}. These challenges are exacerbated in deep architectures prone to overfitting and over-smoothing. To mitigate these issues, several strategies are proposed:
\begin{enumerate}
\item Developing efficient attention mechanisms, such as linear, sparse, and low-rank attention, to reduce complexity and memory usage.
\item Applying graph sparsification or coarsening techniques to decrease graph size and density while preserving key structural and semantic information.
\item Using graph partitioning or sampling methods to divide large graphs into smaller subgraphs or batches for parallel or sequential processing.
\item Exploring graph distillation or compression methods to create compact and effective graph transformer models for deployment on resource-limited devices.
\item Investigating regularization or normalization techniques, such as dropout, graph diffusion, convolution, and graph spectral normalization, to prevent overfitting and over-smoothing.
\end{enumerate}

\subsection{Generalization and Robustness}\label{generalization-and-robustness}

Graph transformers often struggle to generalize to unseen or out-of-distribution graphs, especially those with different sizes, structures, features, and domains. They are also vulnerable to adversarial attacks and noisy inputs, leading to performance drops and misleading results. To enhance the generalization and robustness of graph transformers, consider the following strategies:
\begin{enumerate}
\item Developing adaptive and flexible attention mechanisms, such as dynamic attention, span-adaptive attention, and multi-head attention with different scales, to accommodate varying graphs and tasks.
\item Applying domain adaptation or transfer learning techniques to learn from multiple source domains and transfer knowledge to target domains.
\item Exploring meta-learning or few-shot learning techniques for learning from limited data and rapid adaptation to new tasks.
\item Designing robust and secure attention mechanisms, such as adversarial attention regularization, attention masking, and attention perturbation, to resist adversarial attacks and noisy inputs.
\item Evaluating the uncertainty and reliability of graph transformer models using probabilistic or Bayesian methods, such as variational inference, Monte Carlo dropout, and deep ensembles.
\end{enumerate}

\revtwo{
\subsection{Interpretability and Explainability}\label{interpretability-and-explainability}

Graph transformers, often seen as black box models, face challenges in interpretability and explainability, undermining their credibility and transparency. To address this, several approaches can be considered:
\begin{enumerate}
\item Developing transparent and interpretable attention mechanisms, such as attention visualization, attention attribution, and attention pruning, to highlight the importance and relevance of different nodes and edges in graphs.
\item Applying explainable artificial intelligence (XAI) techniques, such as saliency maps, influence functions, and counterfactual explanations, to analyze and understand the behavior and logic of graph transformer models.
\item Exploring natural language generation techniques, such as template-based generation, neural text generation, and question-answering generation \cite{Guo2023}, to produce natural language explanations for outputs or actions of graph transformer models.
\item Investigating human-in-the-loop methods, such as active learning, interactive learning, and user studies, to incorporate human guidance in the learning or evaluation process of graph transformer models~\cite{elangovan2023effects}.
\end{enumerate}}

\subsection{Learning on Dynamic Graphs}\label{dynamic-and-complex-graphs}
Graphs are dynamic and complex, often changing over time with the addition or removal of nodes and edges, and the modification of their attributes~\cite{febrinanto2023graph,Huang2024}. They may also have diverse types and modalities of nodes and edges. To enable graph transformers to effectively handle such dynamic graphs, consider the following strategies:
\begin{enumerate}
\item Developing temporal and causal attention mechanisms, such as recurrent, temporal, and causal attention, to capture the temporal and causal evolution of graphs \cite{Chen2023}.
\item Applying continual learning techniques on dynamic graphs to avoid forgetting previous knowledge and retraining.
\item Exploring multimodal attention mechanisms, such as image-text, audio-visual, and heterogeneous attention, to integrate multimodal nodes and edges.
\item Leveraging multi-level and multi-layer attention mechanisms, such as node-edge, graph-graph, and hypergraph attention, to aggregate information from different levels and layers of graphs.
\end{enumerate}

\section{Conclusion}\label{conclusion}
Graph transformers are a novel and powerful class of neural network models, which can effectively encode and process graph-structured data. This survey provides a comprehensive overview of graph transformers in terms of design perspectives, taxonomy, applications, and open issues. We first discuss how graph transformers incorporate graph inductive bias, including node positional encoding, edge structural encoding, message-passing bias and attention bias, to encode the structural information of graphs. Then, we introduce the design of graph attention mechanisms, including global and local attention mechanisms. Afterwards, a taxonomy of graph transformers is presented. This survey also includes a design guide for effective graph transformers, including the best practices and recommendations for selecting appropriate components and hyperparameters. Moreover, the application scenarios of graph transformers are reviewed based on various graph-related tasks (e.g., node-level, edge-level, and graph-level tasks), as well as tasks in other domains. Lastly, current challenges and future directions of graph transformers are identified. This survey aims to serve as a valuable reference for researchers and practitioners interested in graph transformers and their applications.


%
%
%

\bibliographystyle{IEEEtran}
\bibliography{IEEEabrv,TNNLS-2024-S-36096}

\vspace{-5mm}
\vskip -2\baselineskip 

\begin{IEEEbiography}[{\includegraphics[width=1in,height=1.25in,clip,keepaspectratio]{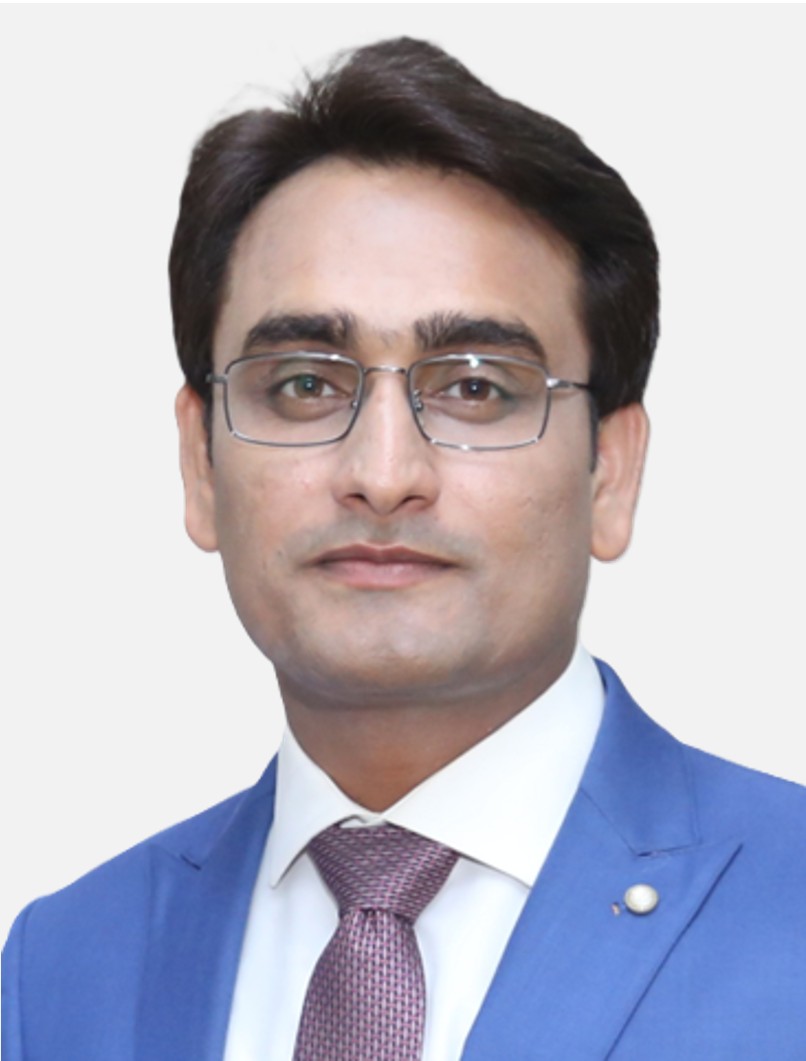}}]{Ahsan Shehzad} received the B.S. degree in computer science from BZU, Multan, Pakistan, in 2015, and the M.S. degree in computer science from Air University, Islamabad, Pakistan, in 2018. He is currently pursuing the Ph.D. degree in software engineering at Dalian University of Technology, China. His current research interests include artificial intelligence, graph learning, health informatics, and brain science.
\end{IEEEbiography}

\vskip -2\baselineskip 

\begin{IEEEbiography}[{\includegraphics[width=1in, clip,keepaspectratio]{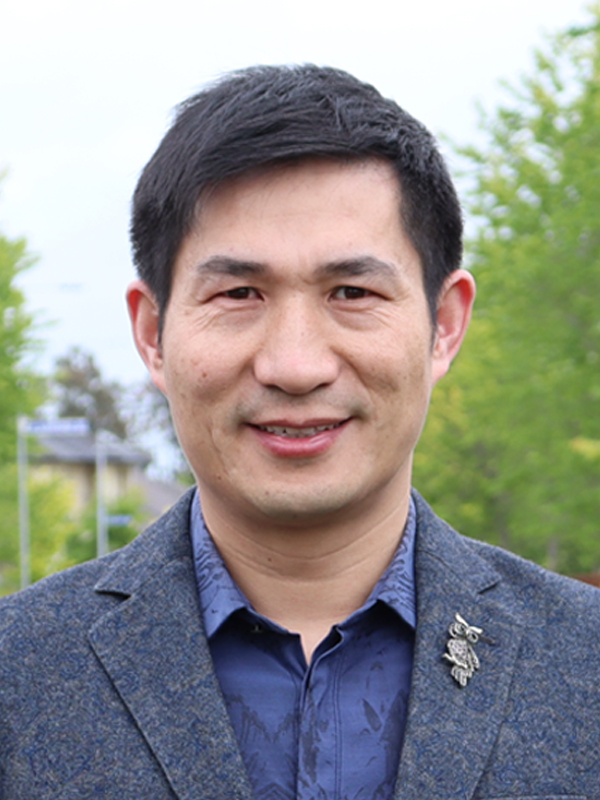}}]{Feng Xia} (Fellow, IEEE) received the BSc and PhD degrees from Zhejiang University, Hangzhou, China. He is a Professor in School of Computing Technologies, RMIT University, Australia. Dr. Xia has published over 300 papers in journals and conferences. He is Chair of IEEE Task Force on Learning for Graphs. His research interests include artificial intelligence, graph learning, brain, robotics, and cyber-physical systems.
\end{IEEEbiography}

\vskip -2\baselineskip 

\begin{IEEEbiography}[{\includegraphics[width=1in,height=1.25in,clip,keepaspectratio]{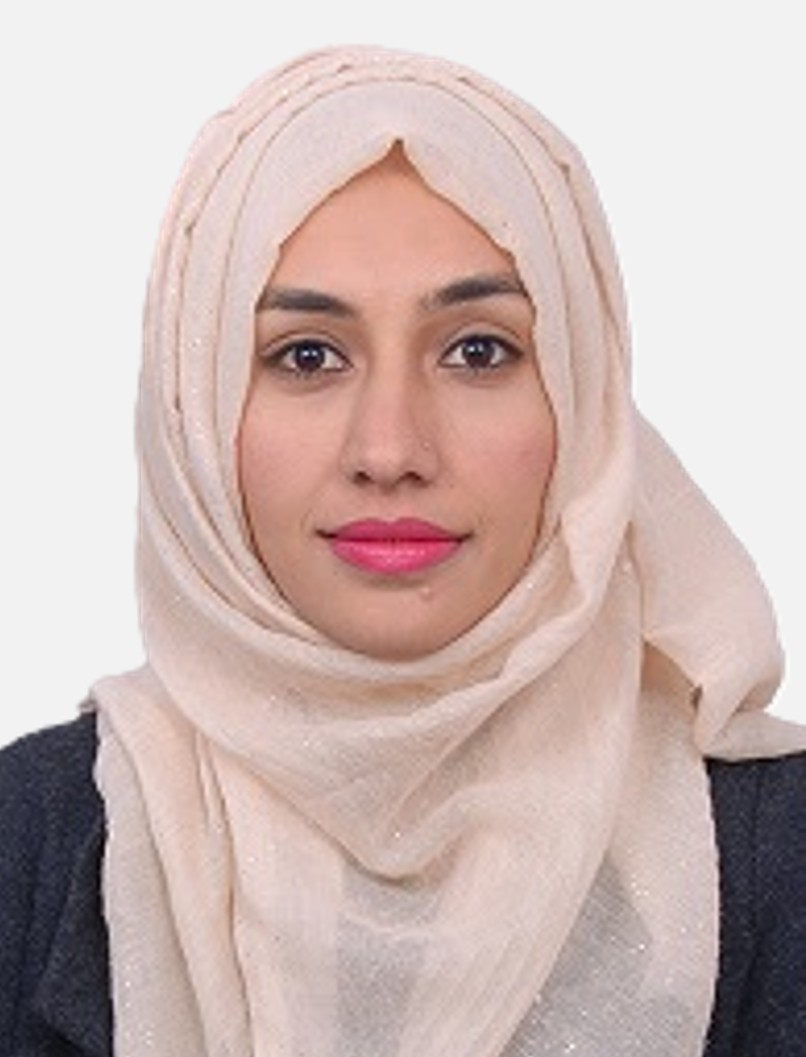}}]{Shagufta Abid} received the B.S. degree in computer science from IUB, Bahawalpur, Pakistan, in 2015, and the M.S. degree in computer science from Air University, Islamabad, Pakistan, in 2019. She is currently pursuing the Ph.D. degree in software engineering at Dalian University of Technology, China. Her current research interests include artificial intelligence, graph learning, and health informatics.
\end{IEEEbiography}

\vskip -2\baselineskip 

\begin{IEEEbiography}[{\includegraphics[width=1in,height=1.25in,clip,keepaspectratio]{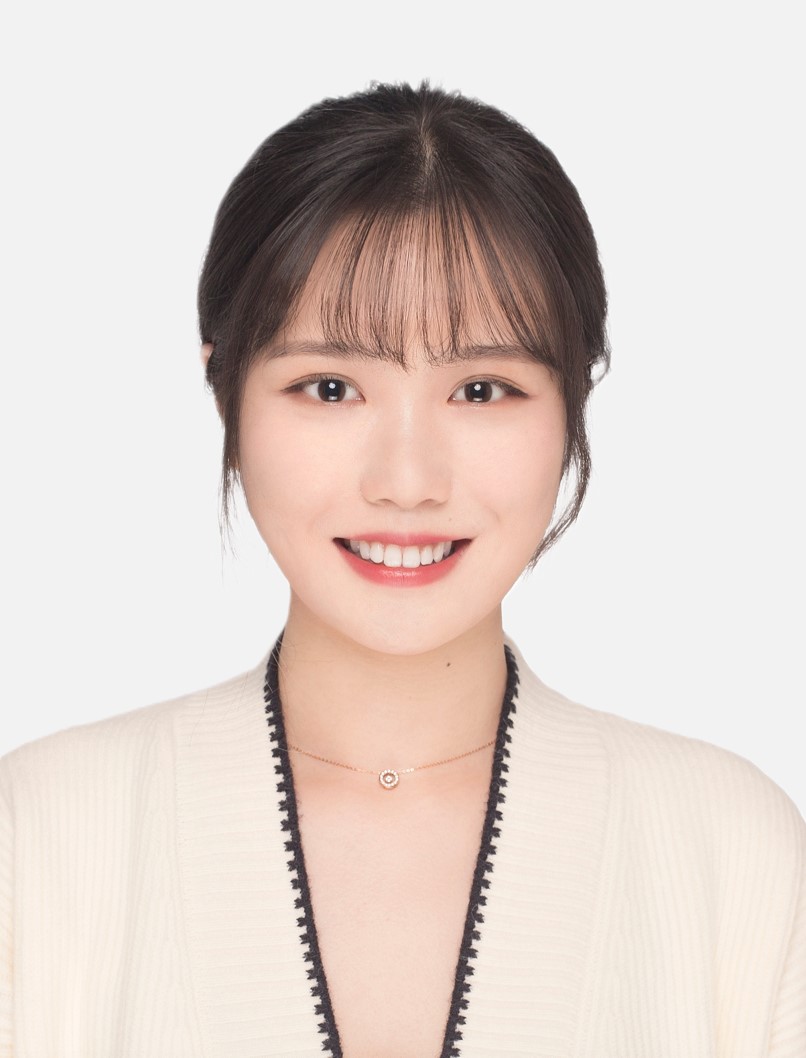}}]{Ciyuan Peng} is a Ph.D. student at the Institute of Innovation, Science and Sustainability, Federation University Australia. She received the B.Sc. degree from Chongqing Normal University, China, in 2018, and the M.Sc. degree from Chung-Ang University, Korea, in 2020. Her research interests include data science, graph learning, brain science and knowledge graphs.
\end{IEEEbiography}

\vskip -2\baselineskip 

\begin{IEEEbiography}[{\includegraphics[width=1in,height=1.25in,clip,keepaspectratio]{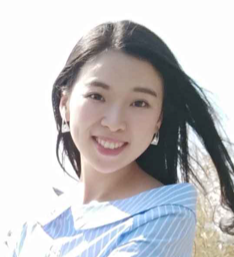}}]{Shuo Yu} (Senior Member, IEEE) received B.Sc. and M.Sc. degrees from the School of Science, Shenyang University of Technology, China. She received a Ph.D. degree from the School of Software, Dalian University of Technology, China. Dr. Shuo Yu is currently an Associate Professor at the School of Computer Science and Technology, Dalian University of Technology. She has published over 50 papers and received several academic awards, including the IEEE DataCom 2017 Best Paper Award, IEEE CSDE 2020 Best Paper Award, and ACM/IEEE JCDL 2020 The Vannevar Bush Best Paper Honorable Mention. Her research interests include data science, graph learning, and knowledge science. 
\end{IEEEbiography}

\vskip -2\baselineskip 

\begin{IEEEbiography}[{\includegraphics[width=1in,height=1.25in,clip,keepaspectratio]{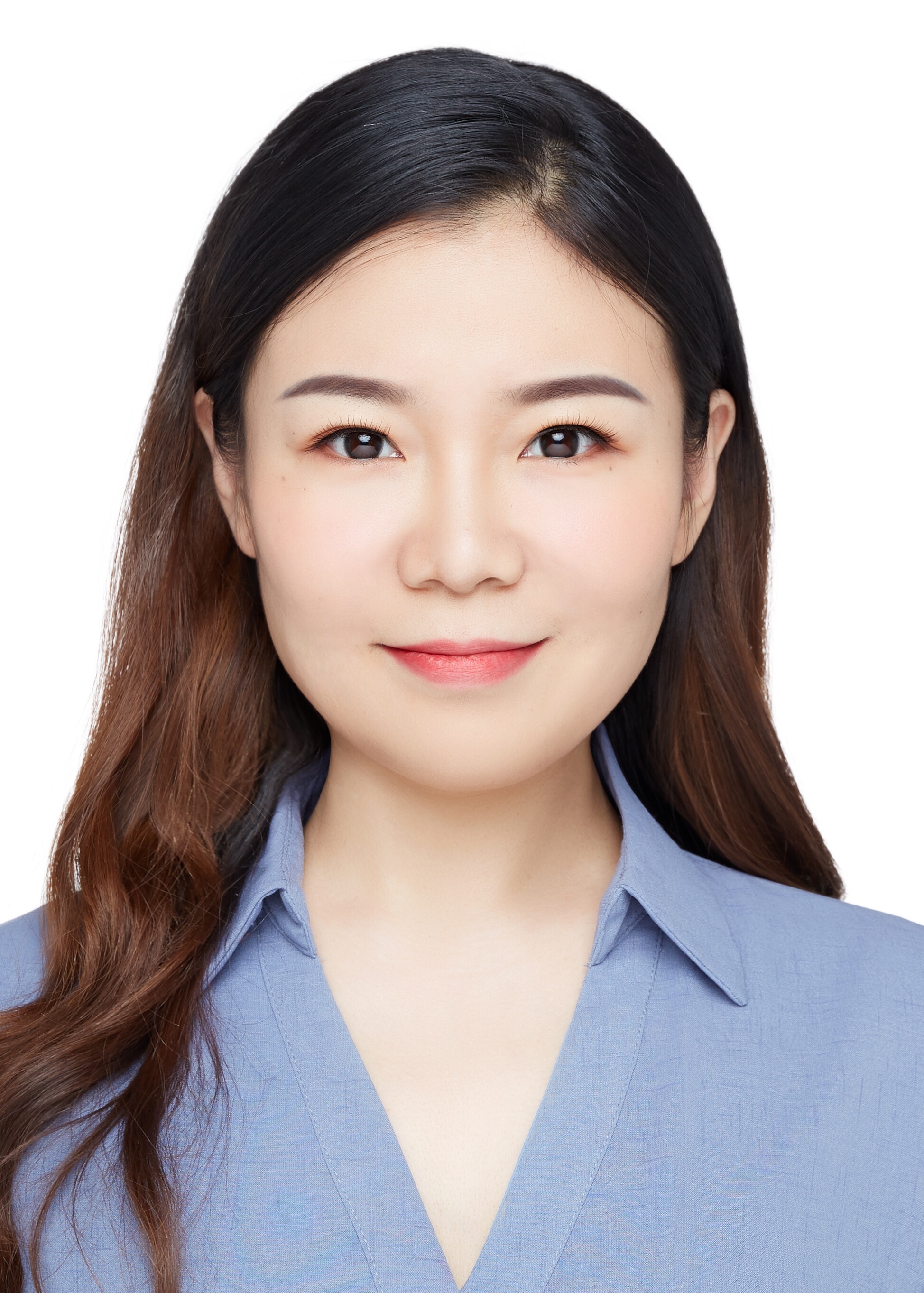}}]{Dongyu Zhang} 	received the MA degree in applied linguistics from Leicester University, UK; and the Ph.D. degree in computer application technology at Dalian University of Technology, Dalian, China. She is currently a full professor in the School of Software at Dalian University of Technology, Dalian, China. Her research interests include natural language processing, sentiment analysis, and social computing. She is a member of the Association for Computational Linguistics, and the China Association of Artificial Intelligence.
\end{IEEEbiography}

\vskip -2\baselineskip 

\begin{IEEEbiography}[{\includegraphics[width=1in,height=1.25in,clip,keepaspectratio]{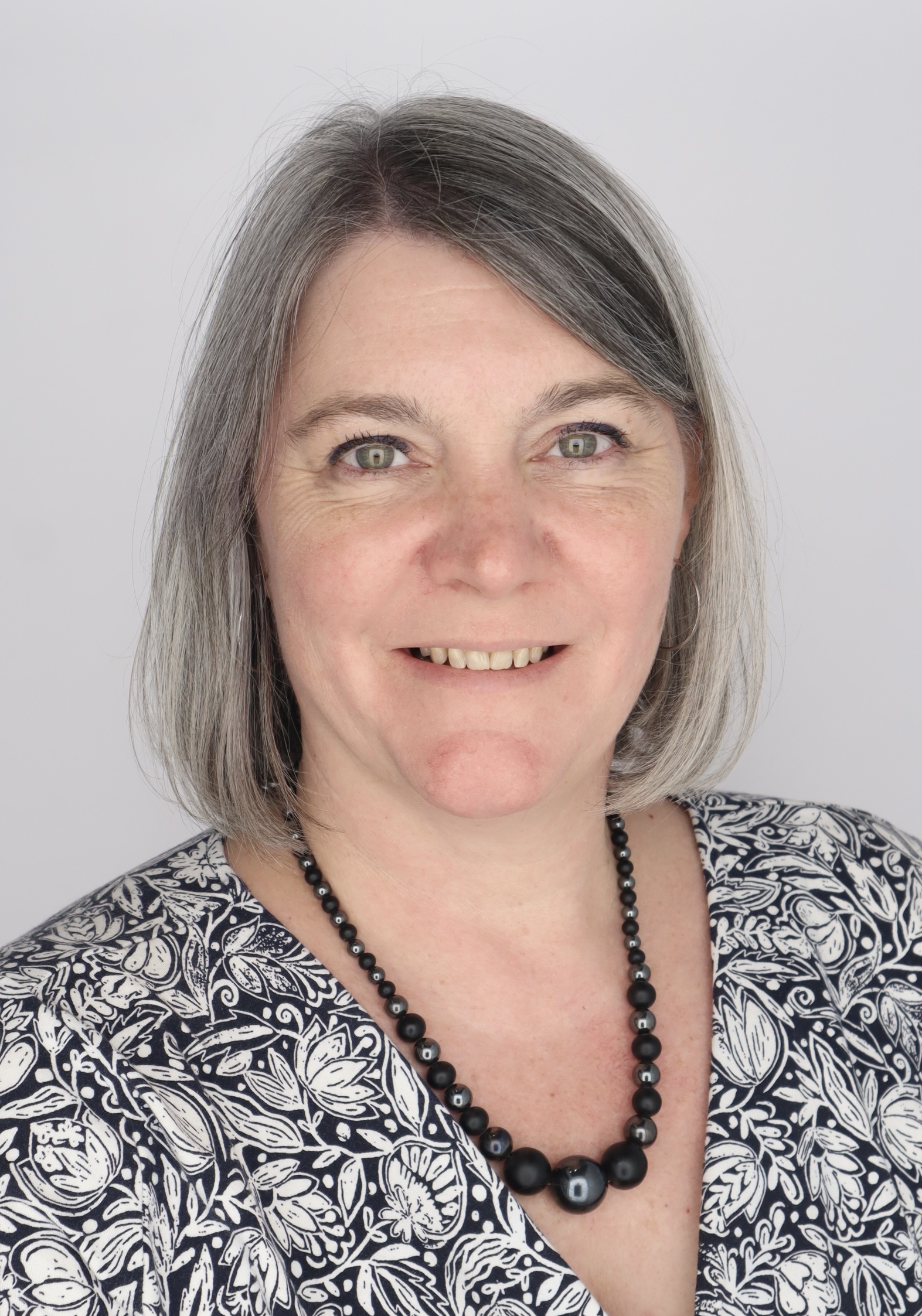}}]{Karin Verspoor} received the BA degree in computer science and cognitive sciences from Rice University (Houston, TX, USA) and the MSc and PhD degrees from the University of Edinburgh (UK). She is currently a Professor and Dean of the School of Computing Technologies, RMIT University in Melbourne, Australia. She has published over 300 papers, with a focus on extraction of information from clinical texts and the biomedical literature with natural language processing and machine learning-based modelling.
\end{IEEEbiography}

\end{document}